%% file: manuscript.tex
\documentclass[twocolumn]{article}

\usepackage{arxiv}
\usepackage[utf8]{inputenc}             
\usepackage[T1]{fontenc}                
\usepackage{url}                        
\usepackage{svg}                        
\usepackage{booktabs}                   
\usepackage{amsfonts}                   
\usepackage{amsmath}                    
\usepackage{braket}                     
\usepackage{nicefrac}                   
\usepackage[letterspace=24]{microtype}  
\usepackage[multiple]{footmisc}         
\usepackage[labelfont=bf]{caption}      
\usepackage{graphicx}                   
\usepackage{cuted}                      
\usepackage[framemethod=tikz]{mdframed} 
\usetikzlibrary{shadows}
\usepackage{tabularx}                   
\usepackage{rotating}                   
\usepackage{chngpage}                   
\usepackage{enumitem}                   
\usepackage[font=small]{caption}        
\usepackage{hyperref}                   

\graphicspath{{./figures/}}             
\definecolor{box-color-inner}{cmyk}{0.38, 0.16, 0.0, 0.0}
\definecolor{box-color-shadow}{cmyk}{0.0, 0.2, 0.0, 0.04}

\mdfsetup{
    shadow              = true,
    shadowsize          = 6pt,
    shadowcolor         = box-color-shadow,
    backgroundcolor     = box-color-inner,
    linewidth           = 0pt,
    rightmargin         = 6pt,
    innertopmargin      = 12pt,
    innerbottommargin   = 12pt,
    innerleftmargin     = 12pt,
    innerrightmargin    = 12pt
}

\hypersetup{
    hidelinks           = true,
    colorlinks          = false, 
    linkcolor           = blue, 
    filecolor           = magenta, 
    urlcolor            = cyan
}

\title{Data Valuation with Gradient Similarity}

\author{ 
    \href{http://orcid.org/0000-0003-2245-8904}{\includegraphics[scale=0.1]{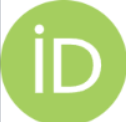}\hspace{1mm}Nathaniel J. Evans} \inst{1},
	\href{https://orcid.org/0000-0002-0144-9614}{\includegraphics[scale=0.1]{icons/orcid.png}\hspace{1mm}Gordon B. Mills} \inst{2,3}, 
     \href{https://orcid.org/0000-0001-8196-1177}{\includegraphics[scale=0.1]{icons/orcid.png}\hspace{1mm}Guanming Wu} \inst{1}, 
	{\hspace{1mm}Xubo Song} \inst{1,3}, 
	\href{https://orcid.org/0000-0001-8333-6607}{\includegraphics[scale=0.1]{icons/orcid.png}\hspace{1mm}Shannon McWeeney} \inst{1,3} 
}

\institute{
    Division of Bioinformatics and Computational Biomedicine, Department of Medical Informatics \& Clinical Epidemiology,
Oregon Health \& Science University, Portland, Oregon, United States of America
    \and
    Division of Oncological Sciences Knight Cancer Institute, Oregon Health \& Science University, Portland, OR, 97201, USA
    \and 
    Knight Cancer Institute, Oregon Health \& Science University, Portland, Oregon, United States of America
}

\contact{evansna@ohsu.edu}

\begin{document}

\twocolumn[
  \begin{@twocolumnfalse}
    \maketitle
    \begin{abstract}
    \input{01_abstract}
    \end{abstract}
    \keywords{Data Valuation \and Deep Learning \and Drug Response \and LINCS}
    \vspace{8pt}
  \end{@twocolumnfalse}
]

\input{02_background}

\input{03_methods}

\input{04_results}

\input{05_discussion}

\section*{Code and Data Availability}

The \href{https://archive.ics.uci.edu/ml/datasets/adult}{Adult}, \href{https://archive.ics.uci.edu/ml/datasets/BlogFeedback}{Blog} and \href{https://www.cs.toronto.edu/~kriz/cifar.html}{Cifar10} datasets can be accessed from the UCI machine learning repository \cite{dua_uci_2019}. The LINCS data can be accessed from the \href{clue.io}{CLUE data library}. All code used for production of the paper figures and the methods described can be found here \url{https://github.com/nathanieljevans/DVGS}. Further questions can be directed to Nathaniel Evans (\url{evansna@ohsu.edu}). 

\section*{Acknowledgements}

This work is supported by the National Library of Medicine (NLM) Training Grant (T15-LM07088). 

The authors thank Dr. Yoon for input on DVRL implementation details and Dr. Ben Cordier for the many discussions about data valuation. 

\bibliographystyle{unsrt}  
\bibliography{references}
\clearpage

\input{06_supplementary}
\clearpage

\end{document}

%% file: 01_abstract.tex
High-quality data is crucial for accurate machine learning and actionable analytics, however, mislabeled or noisy data is a common problem in many domains. Distinguishing low- from high-quality data can be challenging, often requiring expert knowledge and considerable manual intervention. \textit{Data Valuation} algorithms are a class of methods that seek to quantify the \textit{value} of each sample in a dataset based on its contribution or importance to a given predictive task. These data values have shown an impressive ability to identify mislabeled observations, and filtering low-value data can boost machine learning performance. In this work, we present a simple alternative to existing methods, termed \textit{Data Valuation with Gradient Similarity} (DVGS). This approach can be easily applied to any gradient descent learning algorithm, scales well to large datasets, and performs comparably or better than baseline valuation methods for tasks such as corrupted label discovery and noise quantification. We evaluate the DVGS method on tabular, image and RNA expression datasets to show the effectiveness of the method across domains. Our approach has the ability to rapidly and accurately identify low-quality data, which can reduce the need for expert knowledge and manual intervention in data cleaning tasks. 

%% file: 02_background.tex
\section{Background}\label{sec:background}

\subsection{Introduction}\label{sec:intro}

Modern research and "big data" have led to remarkable discoveries and spurred many fields toward high-throughput data collection to capitalize on emerging methods in data science, machine learning, and artificial intelligence. Scientists involved in data collection go to great efforts to generate accurate and reproducible data, however, unavoidable measurement noise, batch effects, and natural stochasticity often lead to varying levels of data quality. Many foundational high-throughput datasets are affected by reproducibility and data quality issues, which often limit the actionable results of these resources \cite{niepel_reproduce, begley_standards, prinz_dti, cheng_systematic_2016}. 

\subsubsection{Data Valuation}

\textit{Data quality} relates to the capacity of data to represent the underlying process. For example, the objective of photography is to gather information about a three-dimensional scene, while the purpose of measuring temperature is to reflect the kinetic energy of an object. Data quality issues can arise from many sources; for instance, chromatic aberration or lens imperfections in photography can distort images, creating inaccurate representations of a scene. Similarly, a miscalibrated thermometer might not measure temperature correctly. Data quality issues can be particularly problematic in machine learning \cite{chen2014review, lukas_mlquality, cai_dataquality}, as a small subset of inaccurate samples can significantly degrade modeling performance even if the majority of samples are high-quality. Curating high-quality datasets can be challenging and usually requires expert knowledge of both the data generation process and the underlying process being measured. A more automated approach to quantify data quality is a class of algorithms called \textit{data valuation}, which assigns a numerical value to each sample in a dataset that characterizes its usefulness toward a predictive task. In the right context, data valuation can effectively capture many aspects of data quality. While there are a number of published data valuation algorithms, many of them follow a similar overarching approach, in which the user must define: 

\begin{itemize} 
    \item \textbf{Source dataset}: The samples that will be valued. Note that this is sometimes called the training dataset\footnote[1]{We use this naming convention to avoid confusion later since DVGS updates model parameters based on gradient from the "Target Dataset" rather than the "Source Dataset." The Data Shapley \cite{ghorbani_data_2019} and Data Valuation with Reinforcement Learning (DVRL) \cite{yoon_data_2019} would refer to this as the "Training" dataset.}.
    \item \textbf{Target dataset}: This dataset characterizes the task or goal of the data valuation, and the choice of alternative target datasets are liable to result in different data values. Note that this is sometimes called the validation dataset \footnote[2]{The Data Shapley \cite{ghorbani_data_2019} and Data Valuation with Reinforcement Learning (DVRL) \cite{yoon_data_2019} would refer to this as the "Validation" dataset.}.
    \item \textbf{Learning algorithm}: The choice of predictive model, e.g., Logistic regression, random forest, neural network, etc. 
    \item \textbf{Performance metric}: The evaluation metric used to compare the learning algorithms predictions against the ground truth, e.g., Accuracy, area-under-the-receiver-operator-curve (for classification), mean-squared-error, $r^2$ (for regression), etc. 
\end{itemize} 

Provided these four user-defined elements, a Data Valuation algorithm then assigns a numerical value to each sample in the source dataset that quantifies the importance of a sample, or its contribution to the predictive performance of the learning algorithm as evaluated on the target dataset. This method can be used in a number of ways, such as:  

\begin{itemize} 
    \item \textbf{Model Enhancement}: To improve the predictive performance of a model by filtering low-quality data or identifying mis-labeled samples. 
    \item \textbf{Attribution}: To quantify data value for monetary recompense or to quantify fair contribution, i.e., credit. 
    \item \textbf{Domain Adaptation}: To identify samples from an alternative domain that are relevant to a target task. 
    \item \textbf{Efficiency}: Reduce the compute resources (run-time or memory) required to train machine learning models. 
\end{itemize}

Existing methods for data valuation include Leave-One-Out (LOO) \cite{cook_detection_1977}, Data Shapley \cite{ghorbani_data_2019}, and Data Valuation using Reinforcement Learning (DVRL) \cite{yoon_data_2019}. Under some conditions, DVRL has been shown to out-perform both Data Shapley and LOO and has been applied to large datasets (more than 500k samples). In noisy or corrupted datasets, these methods can be used to significantly improve machine learning prediction performance by filtering low data values prior to model training. Additionally, data values were shown to effectively quantify data quality aspects such as the amount of noise in an image or incorrect class labels \cite{ghorbani_data_2019} (i.e., low values correlate with high-noise or mislabeled observations). As a demonstration of these methods, a recent paper used Data Shapley to value an x-ray image dataset for the prediction of pneumonia. By removing approximately 20\% of their training data with the lowest data values, the authors were able to improve the test set prediction accuracy by more than 15\%. Furthermore, when the authors inspected a subset of images with the lowest data values, they found it significantly enriched for mislabeled images \cite{tang_data_2021}. 

A key aspect of Data Shapley is the definition of \textit{equitable data conditions} \cite{ghorbani_data_2019}, which we summarize as: 

\begin{itemize} 

    \item{\textbf{Nullity}: If a sample does not affect model performance, it should have a value of zero.}
    \item{\textbf{Equivalency}: Two samples with equal contribution should have equal values.}
    \item{\textbf{Additivity}: The sum of samples data values should be equal to the data value of the grouped samples.}

\end{itemize}

While these conditions are convenient descriptors of data in many settings, they are not required for most of the pragmatic tasks of data valuation. Furthermore, Data Shapley is the only data valuation method to our knowledge with theoretical justifications fulfilling these conditions. Other methods, such as DVRL, perform comparably or better in many data valuation applications, such as corrupted label identification \cite{yoon_data_2019}. 

\subsection{Library of Integrated Network-Based Cellular Signatures}

There are few, if any, datasets devoid of data quality issues, and addressing these challenges can improve the results of downstream analytics. A foundational dataset that has been highly impactful in modern research, especially in the cancer and drug-development domain, is the Library of Integrated Network-Based Cellular Signatures (LINCS) project. The LINCS program has generated high-dimension transcriptomic profiles (L1000 assay; 978 \textit{landmark} genes) characterizing the effect of chemical and genetic perturbations across a range of cellular contexts, time points, and dosages \cite{subramanian_lincs}. This data has been used successfully in many applications; however, a continued challenge with high-throughput data pipelines is the identification of low-quality samples. In 2016, a systematic quality control analysis of LINCS L1000 data showed that differentially expressed genes (DEGs) inferred from the L1000 platform were often unreliable. For example, only 30\% of DEGs overlapped between any two selected control viral vectors in short-hairpin RNA (shRNA) perturbations \cite{cheng_systematic_2016}. To address these issues, many researchers have proposed methods to improve the L1000 data analysis pipeline, including alternative approaches to peak deconvolution \cite{yue_peak1, li_peak2}, and a novel method of aggregating bio-replicates in order to improve the noise-to-signal ratio \cite{clark_cdir, duan_l1000}. 

A recent paper, which sought to use the LINCS L1000 dataset for the repurposing of COVID-19 drugs, proposed a simple but effective method of quantifying sample-level data quality by computing the average Pearson correlation (APC) between the replicates of a perturbation. Intuitively, if replicates are discordant, and therefore have low or negative pairwise correlations, then the resulting APC value is low; however, if the replicates are concordant and have high pairwise correlations, then the APC value is high. The authors went on to show that filtering L1000 data based on APC values could significantly improve the predictive accuracy of machine learning models \cite{pham_deep_2021}. 

Improvement of data quality in large publicly available datasets, such as the LINCS project, has the potential to markedly improve the usefulness and impact of these datasets. In addition, effective data quality metrics could be used to inform the selection of new conditions that will be most beneficial to select prediction tasks or to avoid conditions that are unlikely to be useful.  

\subsection{Related Work}\label{sec:related_work}

\textit{Dataset Distillation} is a related field, which attempts to distill knowledge from a large dataset into a small one by synthesizing a new dataset that is representative of the original dataset but much smaller \cite{zhu_dataset_distillation, yu_dataset_distillation}. Adjacent to this domain is \textit{core-set} or \textit{instance selection} that focus on selecting a subset of a dataset that leads to comparable or better machine learning performance. In many pragmatic applications, \textit{data valuation} can be seen as coreset or instance selection method; For instance, data valuation produces a ranked list of the samples in a given dataset, based on their value or usefulness towards a predictive task. A ranked list of observations can easily be treated as an instance selection problem by choice of a threshold. Selection of a data value threshold, either by post-hoc analysis or manual choice, reframes data valuation methods as a \textit{ instance selection} approach. Additionally, many of the evaluation techniques of common data valuation methods are analogous to instance selection  (e.g., machine learning performance improvement goals). There is no analog for the equitable data value conditions described by Ghorbani et al. \cite{ghorbani_data_2019} in core-set or instance selection. Several notable methods of core-set or instance selection includes \textit{herding} \cite{welling_herding, welling_herding2}, distribution-matching \cite{bach_np_coresets, feldman_mm_coreset} and incremental-gradient matching approaches \cite{mirzasoleiman_craig}. There have also been instance selection approaches for large language models, which require large amounts of data to train, and the choice of prompting can have drastic impacts on model performance \cite{lu_promptpg, xie_dsir}. 

\textit{Anomaly detection} or \textit{outlier detection} attempts to separate data instances that deviate from the majority of samples \cite{Pang_2021}. Data valuation, especially when used to identify corrupted labels or characterizing exogenous feature noise, can be examined from the lens of anomaly detection. For instance, the DVRL \textit{Estimator} model tries to learn a joint probability distribution of exogenous and endogenous features that maximizes predictive performance of a given learning algorithm. If we make the assumption that identifying in-distribution training data will lead to test performance generalization, then DVRL can be thought of as a method for separating anomalous (out-of-distribution) from normal samples (in-distribution). There have been countless methods introduced for anomaly detection, however, of particular relevance to this paper is a gradient-based anomaly representation for autoencoders proposed by Kwon et. al, which defines an anomaly score based on both reconstruction error and the gradient. \cite{kwon_gradcon}.

There has also been significant research on how to train machine learning models in the presence of noisy or corrupted data. These methods range broadly and include meta learning, sample re-weighting schemes \cite{ren_reweight, jiang_mentornet}, noise-robust loss functions \cite{zhang_lossfn} and loss correction algorithms \cite{hendrycks_trusted}. These methods predominately focus on training high-performing models without explicitly removing corrupted or spurious observations; however, several of these methods use re-weighting schemes that rely on interim observation-specific weights and could be considered analogous to data values.

\subsection{Contributions}\label{contributions}

Data valuation is an efficient and automated approach to characterizing sample informativeness, particularly in data cleaning tasks such as identifying incorrectly labeled or noisy samples. Existing data valuation methods, however, have limitations that hinder widespread application. Data Shapley does not scale well to large datasets and underperforms in certain tasks like corrupted label identification compared to DVRL. DVRL often exhibits high performance in data valuation applications, but is sensitive to hyperparameters, choice of dataset, and predictive model. It can be inconvenient and time consuming to tune the DVRL hyperparameters and is ineffective in some predictive tasks. Furthermore, while DVRL is significantly faster than Data Shapley, this method still requires sequential training of models to accurately estimate data values, which consumes significant computational resources.

In this paper, we introduce a novel data valuation method and compare it against baselines in two key tasks: 1) identifying corrupted labels and 2) identifying samples with high exogenous feature noise. We also explore the application of data valuation in unsupervised learning settings, which to our knowledge is the first method to evaluate this. Unsupervised data valuation is ideal for quantifying sample noise in biological data types such as 'omics sequencing data (RNA expression, DNA mutation, methylation, etc.). Finally, we apply our method to compute data values for the LINCS L1000 level 5 dataset, which contains more than 700,000 high-dimensional samples. Our method demonstrates performance comparable to that of Data Shapley \cite{ghorbani_data_2019} and DVRL \cite{yoon_data_2019} while being significantly more computationally efficient. The speed and scalability of our method make it applicable to large datasets, even with small compute budgets. Moreover, our method is robust to hyperparameters, making it user-friendly.

Although data quality metrics have been proposed for the LINCS L1000 dataset, such as the average Pearson correlation (APC) between replicates \cite{pham_deep_2021}, our data valuation results offer an alternative data quality metric. We show that filtering data based on our data values results in equivalent or higher-performing models than data filtering based on APC. Additionally, we show that our method is more effective in capturing high-valued samples than the APC metric, which could be used to inform future data acquisition decisions.

%% file: 03_methods.tex
\section{Proposed Methods}\label{sec:methods}

\subsection{Data Valuation with Gradient Similarity}

\begin{figure}[t]
\centering
\captionsetup{width=\linewidth}
\includegraphics[width=\linewidth]{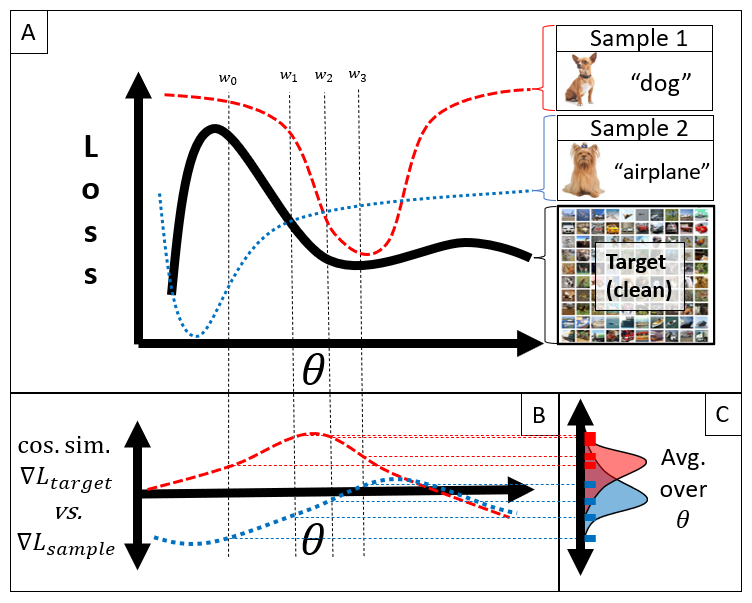}
\caption{We propose a method of data valuation that compares each source sample to the target samples by computing the similarity of gradients during stochastic gradient descent. In panel A, we depict a toy-example of a 1-d loss landscape. Sample 1 (red) is an accurately labeled (high-quality), whereas sample 2 (blue) is incorrectly labeled (low quality). In panel B, we plot the similarity of each source sample gradient compared to the target set gradient (black solid line in panel A). Panel C shows the marginal distribution of gradient similarities, which is averaged to obtain the final source sample data value. To make this process tractable, gradient similarities are computed over a limited number of model parameter values during traditional stochastic gradient descent. The computed gradients are visualized by dotted lines in panels A,B and C ($w_0$, $w_1$,...,$w_3$). To choose the relevant values of $\theta$, we use stochastic gradient descent (SGD), with gradients calculated from the target set.}
\label{fig:dvgs_overview}
\end{figure}

We propose a method of Data Valuation with Gradient Similarity (DVGS), based on the premise that \textbf{source samples with a loss surface similar to the target loss surface will be more useful to a shared predictive task than source samples with dissimilar loss surfaces}. For instance, a training dataset loss surface with a similar shape and minima to the validation dataset loss surface is likely to positively contribute to the validation predictive task. This premise is visualized by a toy example in Figure \ref{fig:dvgs_overview}. Analytically computing the loss criteria for all possible parameter values (i.e., the full loss surface) is intractable for most problems, and therefore a comprehensive comparison of loss surfaces is challenging. However, we can approximate the comparison of loss surfaces by comparing gradient similarities at select parameter values. Comparison of gradients is also advantageous as it factors out the absolute loss value.

Similarly to other data valuation methods, DVGS requires a target dataset that characterizes the desired predictive task. The target dataset may be of high quality, specific prediction domain, or a randomly sampled holdout set. Additionally, the user must define a differentiable predictive model that can be trained using stochastic gradient descent (SGD). The source dataset serves as input on which data valuation will occur, with the goal of characterizing useful or detrimental samples. To perform DVGS, we optimize model parameters using SGD on the target dataset and at each iteration compute the similarity of the target batch gradient to each source sample gradient. We posit that this approach will accurately estimate data values if the gradient similarities are measured in critical regions of the weight-space, such as regions commonly explored during optimization. This procedure is documented in Algorithm 1. We do not expect or justify that this approach satisfies the equitable data value conditions proposed by Ghorbani et al., however, we empirically demonstrate that this approach effectively characterizes data quality in many real-world prediction tasks while being simple, scalable, and easily extensible to a wide range of model architectures and predictive tasks. 

\begin{algorithm*}[t]
\centering 
\caption{Data Valuation with Gradient Similarity} 
\begin{algorithmic}[1]

\Require {Differentiable model ($f_{\theta}$), learning rate ($\alpha$), source dataset ($\mathcal{D}_s$), target dataset ($\mathcal{D}_t$), number of training iterations ($N_{iter}$), target batch size ($R$), loss criteria ($\mathcal{L}$), and similarity criteria ($C$). \newline}

\For {$i=0,1, \ldots, N_{iter}$}

    \State $B_i \sim \mathcal{D}_{t} $ \Comment{sample mini-batch from target set}
    \For {$j=0,1,\ldots,R$}
        \State $x_j, y_j \sim B_i$
        \State $\hat{y_j} \leftarrow f_{\theta}(x_j)$ \Comment{predict outcome for target batch}
    \EndFor
    \State $\nabla \mathcal{L}^{target}_i \leftarrow \frac{\partial}{\partial \theta}(\frac{1}{R}\sum_{j=0}^{R}\mathcal{L}(\hat{y}_j, y_j))$ \Comment{compute target batch gradient}
    \For {$k=0,1,\ldots,N_{source}$}
        \State $x_k, y_k \sim \mathcal{D}_s$
        \State $\hat{y_k} \leftarrow f_{\theta}(x_k) $ \Comment{predict outcome for source sample}
        \State $\nabla \mathcal{L}^{source}_k \leftarrow \frac{\partial}{\partial\theta}(\mathcal{L}(\hat{y}_k, y_k))$ \Comment{compute the gradient for the source sample}
        \State $\nu_k^i \leftarrow C(\nabla \mathcal{L}^{source}_k, \nabla \mathcal{L}^{target}_i)$ \Comment{compute similarity of source sample gradient to the target batch gradient}
    \EndFor
    \State $\theta_{i+1} \leftarrow \theta_i - \alpha \nabla \mathcal{L}^{target}_i$ \Comment{update model parameters using the target batch gradient}
\EndFor
\For {$k=0,1,\ldots,N_{source}$}
    \State $\nu_k \leftarrow \frac{1}{N_{iter}} \sum_{i=0}^{N_{iter}}{\nu_k^i}$ \Comment{compute the average gradient similarity for each source sample}
\EndFor
\end{algorithmic} 
\end{algorithm*}

Calculating the similarity between the gradients of the source samples and the target dataset requires a function that takes as input two high-dimensional gradient vectors and returns a single scalar characterizing similarity. Theoretically, any distance metric is applicable here, however, we chose to use cosine similarity because it produces easily interpreted values between [-1,1] and neglects vector magnitude. We were concerned that gradient magnitudes may vary between early- and late-stage training, and to avoid biasing data values by large gradient magnitudes, we rationalize that gradient magnitude should be ignored.

In classification problems, each class is likely to induce a distinct gradient, and therefore target sets with a class imbalance are likely to introduce class-specific biases to data values. For instance, in a binary classification problem, if the target set has a majority of the positive class, then the source samples with the negative class may be particularly dissimilar, even if they are valuable to the optimization process. To avoid inadvertent bias of class-based data values, we suggest balancing class weights \cite{zeng_classweights} when computing target gradients. Future approaches should explore the comparison of within-class gradient similarities, which may mitigate this problem without class balancing. 

Intuitively, the choice of initialization weights is likely to produce different data values, especially if the target set has a complex multimodal loss surface. To prevent variance in DVGS data values due to weight initialization or stochastic mini-batch sampling, we add the option to run the DVGS algorithm multiple times, each with unique weight initialization and randomization seeds. Using this approach enables DVGS to explore multiple minima and compute similarity values on a wider range of parameter values. To aggregate a final data value, gradient similarities are averaged across all iterations and runs. 

\subsection{Time Complexity}
In most applications, it is reasonable to assume that the target dataset is much smaller than the source dataset, and therefore most of the runtime is spent computing the source gradients. This can be partially mitigated by only computing gradient similarities every $T$ iterations or by pretraining the model. We estimate\footnote{See supplementary note \ref{supp:time} for experimental evaluation of time complexity.} the computational complexity in big O notation: 

$$ \mathcal{O}(\frac{N_{iter} N_{source}}{T}) $$

We expect that the DVGS method will scale linearly with the number of source samples and training iterations. A particular advantage of the DVGS methods is that only a single model need be trained, whereas Data Shapley and DVRL require training many models sequentially. This time complexity makes it suitable for application to large datasets. Additionally, DVGS can be run in parallel and the results averaged to compute more accurate data values; Such an ensemble approach is ideal for large datasets and complex loss surfaces. In many tasks, such as image classification with convolutional neural networks, it can be advantageous to pretrain the convolutional layers prior to performing DVGS.

 \subsection{Data} 

 In this paper, we apply our data valuation algorithm to four datasets under various conditions. 

 \begin{itemize} 
    \item The \textbf{ADULT} dataset, also known as the "census income" dataset, consists of 14 categorical or integer features representative of an adult individual and labeled based on whether they make more than 50k dollars per year \cite{dua_uci_2019}.
    \item The \textbf{BLOG} dataset consists of internet blog characteristics parsed from the raw HTML file and the output is the average number of comments received; We then binarize the endogenous variable with threshold of 0 \cite{kriszti_blog}. 
    \item The \textbf{CIFAR10} dataset, which consists of tiny images labeled as one of 10 possible objects \cite{krizhevsky_cifar10}; we transform the images into an informative feature representations using a pre-trained InceptionNet prior to data valuation \cite{szegedy_inceptionNet}.
    \item The \textbf{LINCS L1000} dataset measures RNA expression in cell lines some time after a chemical or genetic perturbation \cite{subramanian_lincs} We further break the LINCS L1000 into two data partitions: 1) all data and 2) high-APC (>0.5) data (see supp. note \ref{supp:apc}). 
\end{itemize}

We chose the first three datasets and pre-processing steps (ADULT, BLOG, and CIFAR10) to match the evaluations performed in previous work \cite{yoon_data_2019, ghorbani_data_2019}. Similarly, we try to match the respective dataset size (target, source, test) choices made in previous work to provide similar evaluations. 

The LINCS L1000 is a widely used biological dataset that suffers from known data quality issues \cite{subramanian_lincs, duan_l1000, li_peak2, yue_peak1, niepel_reproduce, pham_deep_2021} and removing inaccurate or noisy samples from this dataset could benefit the cancer drug response domain. 

\subsection{Dataset Corruption} 

To simulate poor data quality, we artificially corrupt datasets in two ways:

\begin{itemize} 
    \item{\textbf{Label Corruption}; Endogenous variable (y)} 
    \item{\textbf{Feature Corruption}; Exogenous variable (x)} 
\end{itemize} 

Labels are corrupted by randomly relabeling a proportion of the source dataset class labels; for instance, an image of a "dog" might be re-labeled as "cat". The corrupted sample indices are then used as the ground truth of data quality and can be compared to data values. The expectation is that corrupted labels will have lower data values indicating that they are less valuable to model performance. To summarize the ability of data values to identify corrupted samples, we use the area under the receiver operator curve (AUROC) metric: 

$$ AUROC(c, -\nu) $$

Where $c$ is the corrupted label mask $(0=uncorrupted; \ 1=corrupted)$ and $\nu$ is the data values. Notably, we flip the data value sign as we expect large data values to indicate high quality data, and small data values to indicate low quality or mislabeled observations.  

To explore the ability of data valuation to capture exogenous feature sample quality, we add Gaussian noise to each observation: $$ x_{i,j}^* = \mathcal{N}(0, \phi_i) + x_{i,j} $$ where $x_{i,j}^*$ is feature $j$ of the corrupted sample $i$, and $\phi_i$ is an observation-specific noise rate sampled from a uniform distribution. Thus, samples with larger noise rates ($\phi_i$), will have noise with greater variance. The primary evaluation task is to apply data valuation and compare the data values with the sample-specific noise rates. We expect that samples with large noise rates will have small data values, indicating that they are less valuable to model performance. To evaluate performance on this task, we use Spearman correlation \cite{Spearman1987ThePA}. Note that we change the sign of our data values as we expect that high data values should correlate with large noise rates: 
$$ \rho = Spearman(\phi, -\nu) $$

%% file: 04_results.tex
\section{Results}\label{sec:results}

\subsection{Label Corruption}\label{label corruption}

To evaluate the ability of data values to capture mislabeled samples, we artificially corrupt labels in three classification datasets: ADULT, BLOG, and CIFAR10. We compare DVGS to several baseline methods:

\begin{itemize}
    \item {Randomly assigned data values (null model)}
    \item{Leave-out-out (LOO) \cite{cook_detection_1977}}
    \item{Truncated Monte-Carlo Data Shapley (dshap) \cite{ghorbani_data_2019}}
    \item{Data Valuation with Reinforcement learning (DVRL) \cite{yoon_data_2019}}
\end{itemize}

The Leave-one-out and Data Shapley algorithms are only applied to the ADULT and BLOG datasets due to compute resource constraints.

In all three datasets, we corrupt 20\% of the labels. For the ADULT and BLOG datasets we use 1000 source observations and 400 target observations. For the CIFAR10 dataset, we use 5000 source observations and 2000 target observations. We expect accurate data valuation to produce values such that corrupted samples data values will be smaller than uncorrupted samples, indicating that they are less valuable or useful toward our target predictive task. Additionally, we expect that filtering corrupted labels should improve model performance. In each experiment, we evaluate the ability of data values to 1) identify corrupted labels and 2) modify model performance as measured on a hold-out test set when we filter a proportion of the dataset. In this second task, we evaluate the performance changes when we filter high-values (expectation that performance will decrease) versus low-values (expectation that performance will improve or be unaffected).

\begin{figure*}[ht!]
     \centering
     \begin{subfigure}{0.3\textwidth}
         \centering
         \includegraphics[width=\textwidth]{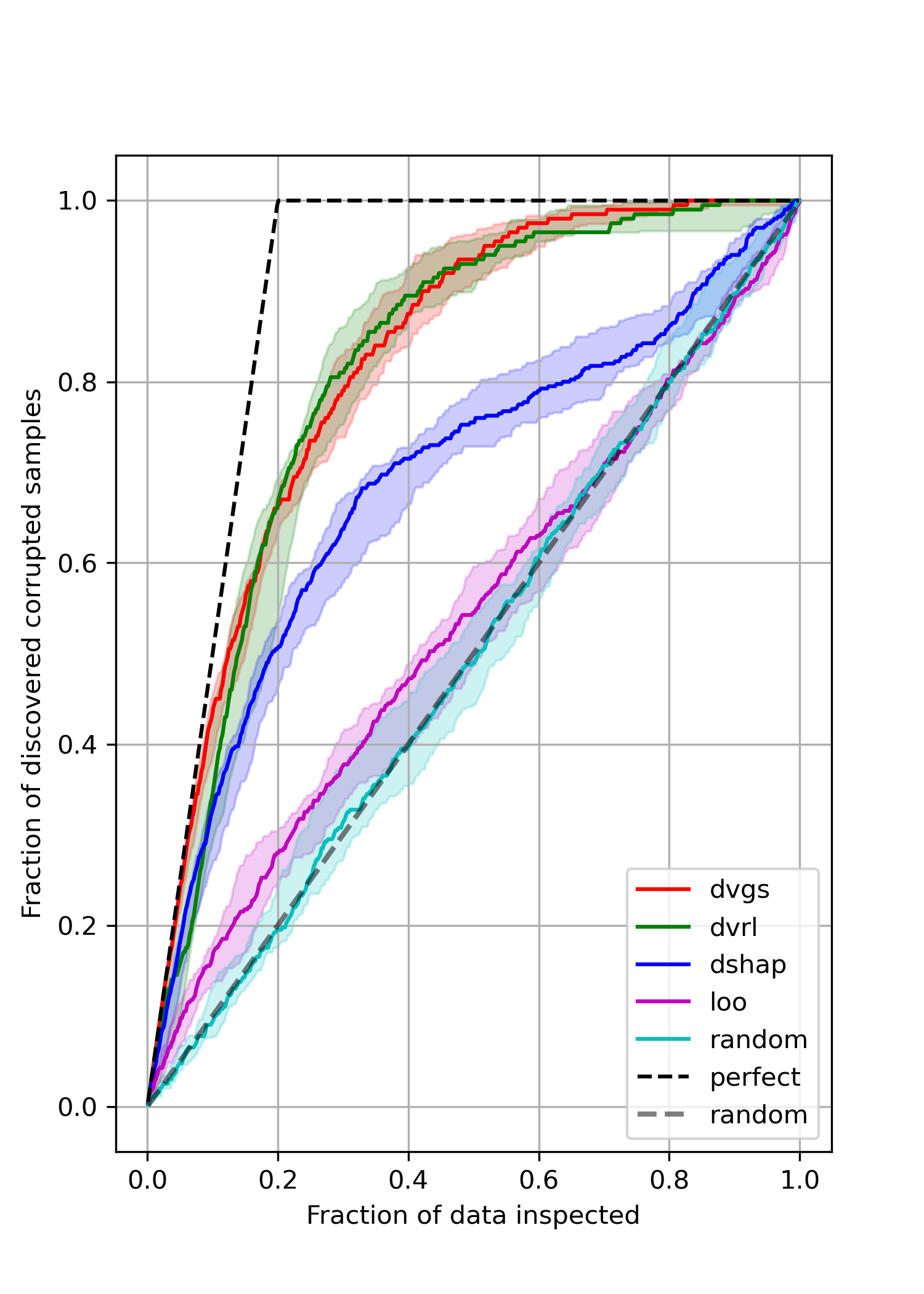}
         \caption{ADULT Dataset}
         \label{fig:exp1_labels}
     \end{subfigure}
     \hfill
     \begin{subfigure}{0.3\textwidth}
         \centering
         \includegraphics[width=\textwidth]{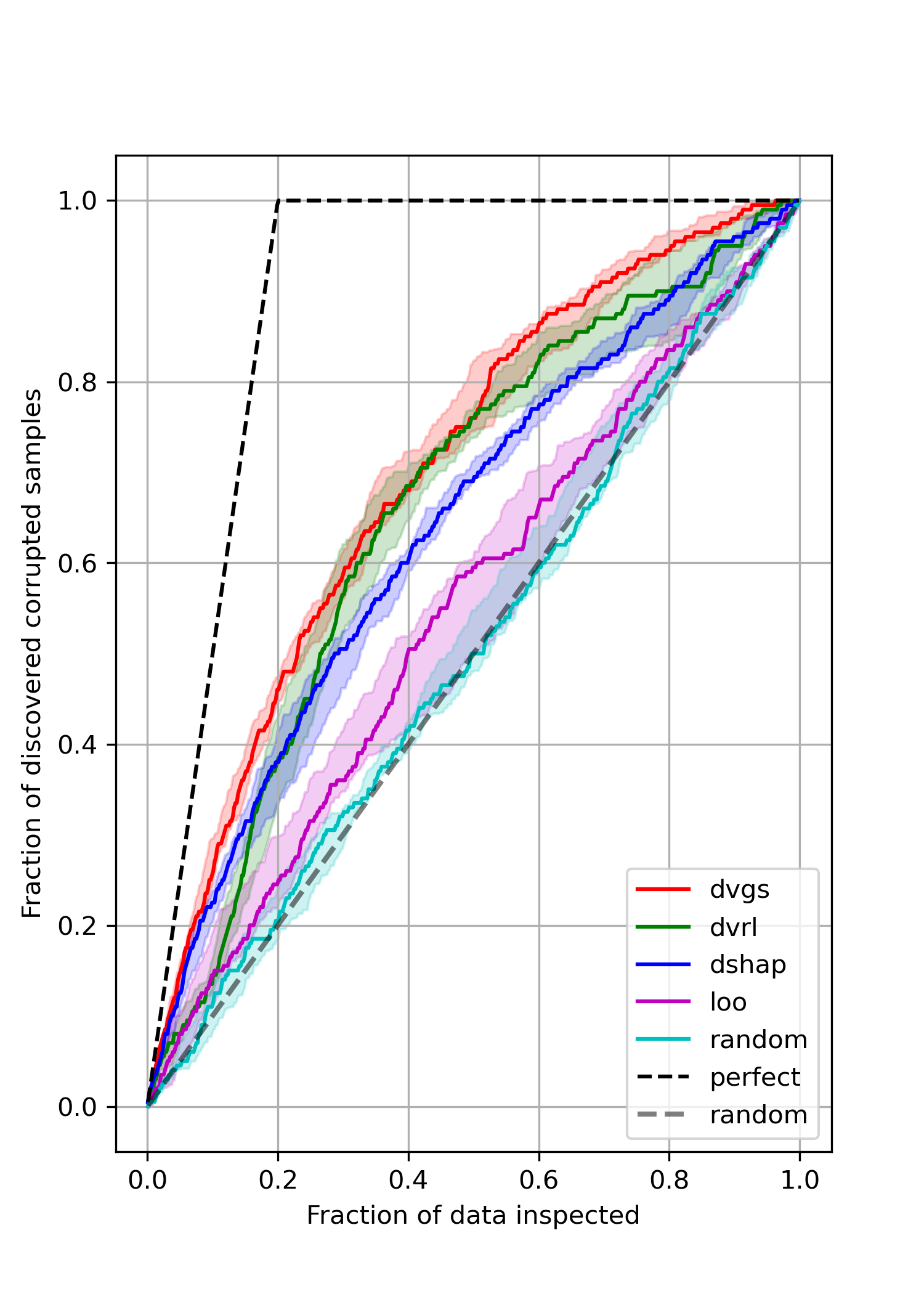}
         \caption{BLOG Dataset}
         \label{fig:exp2_labels}
     \end{subfigure}
     \hfill
     \begin{subfigure}{0.3\textwidth}
         \centering
         \includegraphics[width=\textwidth]{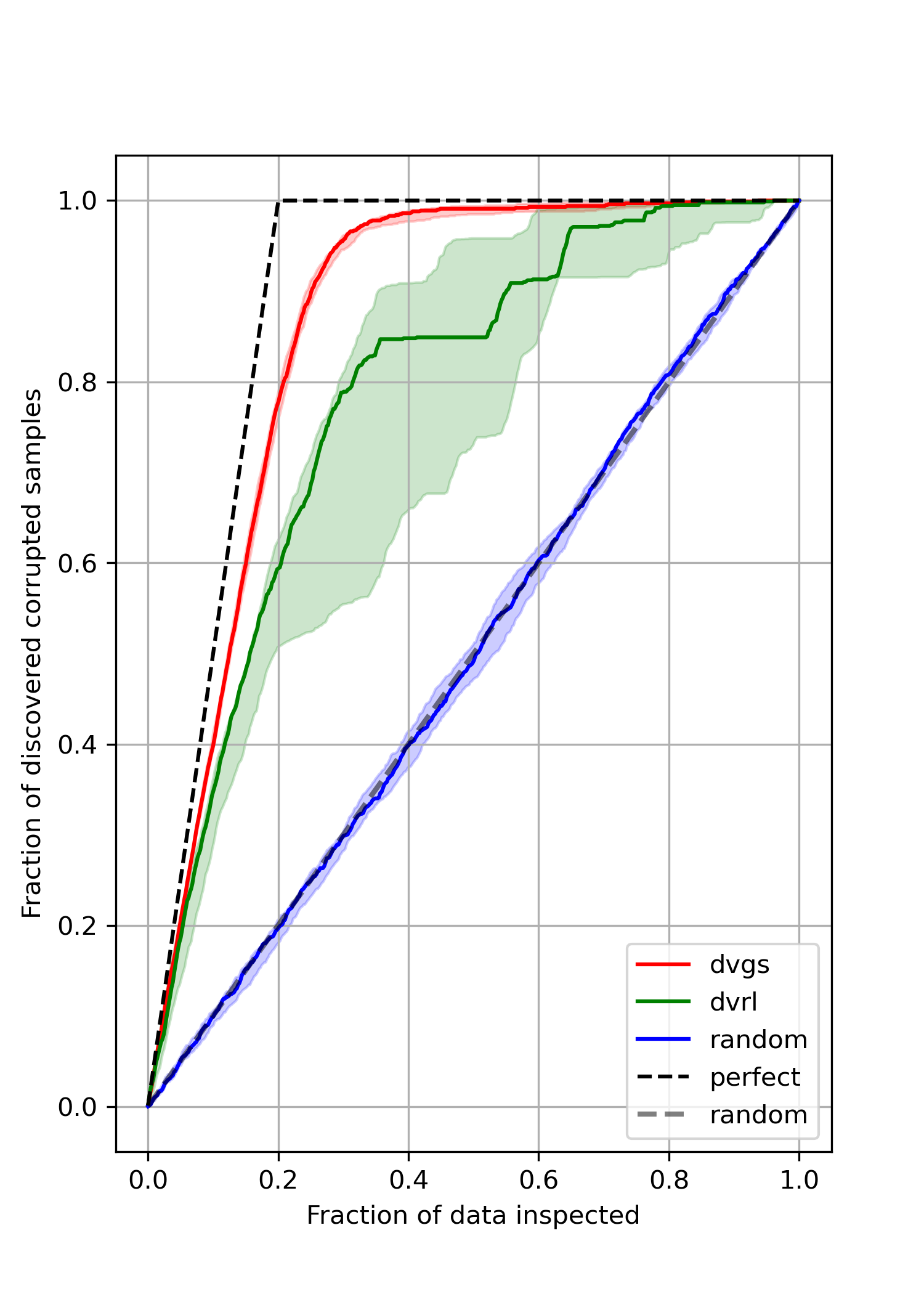}
         \caption{CIFAR10}
         \label{fig:exp3_labels}
     \end{subfigure}
\caption{Evaluation of respective data valuation methods ability to identify corrupted labels. The Gray dashed "random" are theoretical random performance, whereas blue/cyan "random" is empirically measured random values.}
\label{fig:exp123_labels}
\end{figure*}

\begin{figure*}[ht!]
     \centering
     \begin{subfigure}{0.3\textwidth}
         \centering
         \includegraphics[width=\textwidth]{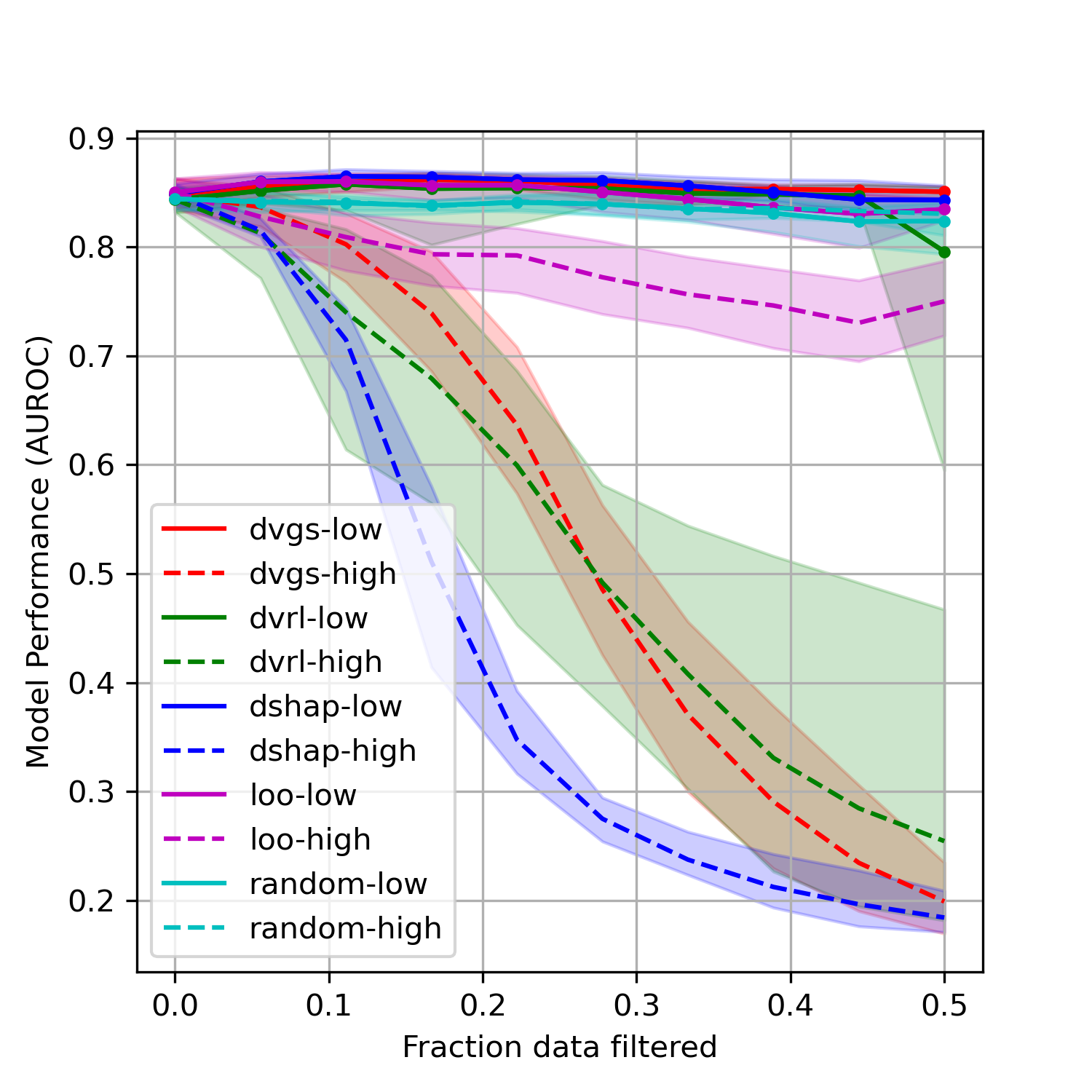}
         \caption{ADULT Dataset}
         \label{fig:exp1_perf}
     \end{subfigure}
     \hfill
     \begin{subfigure}{0.3\textwidth}
         \centering
         \includegraphics[width=\textwidth]{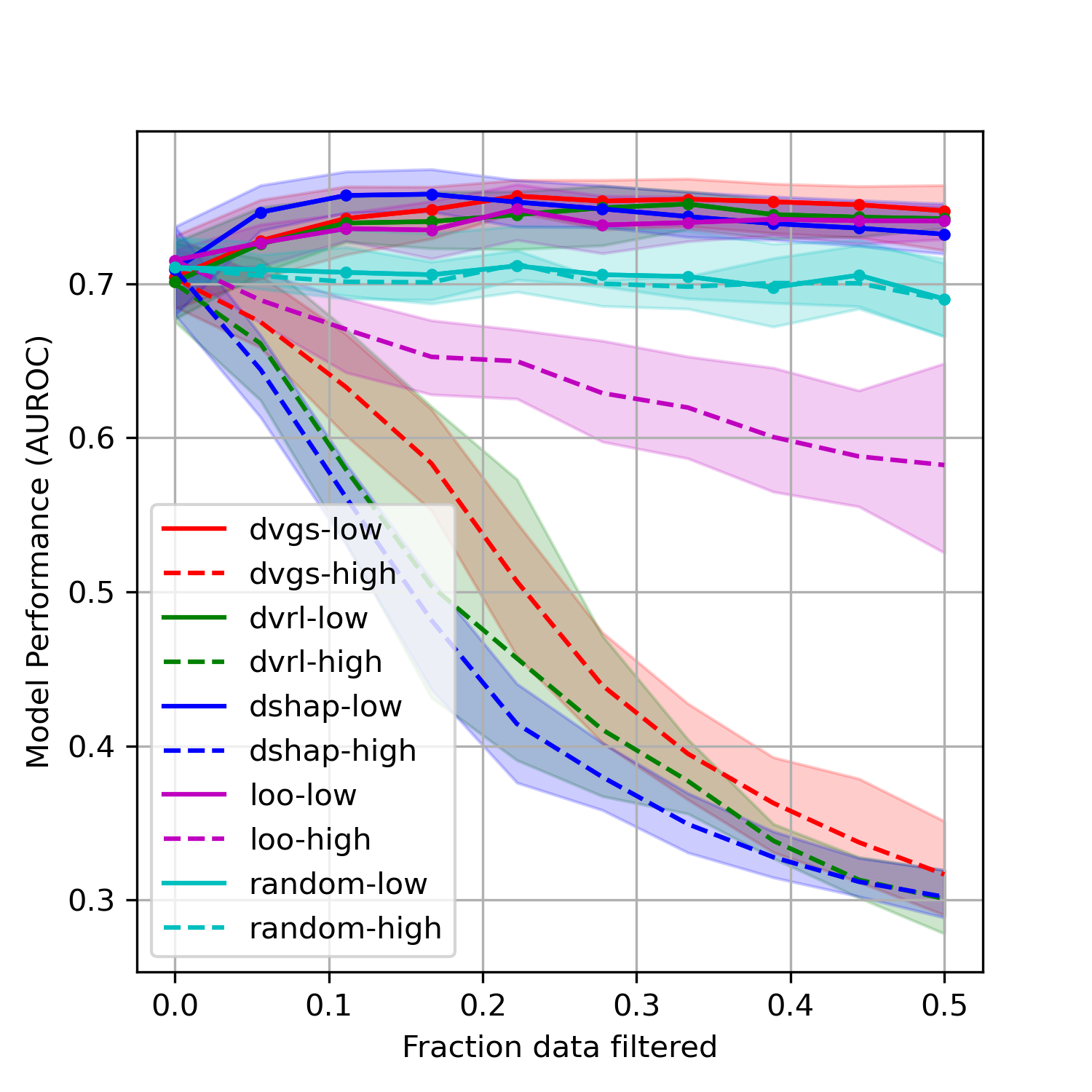}
         \caption{BLOG Dataset}
         \label{fig:exp2_perf}
     \end{subfigure}
     \hfill
     \begin{subfigure}{0.3\textwidth}
         \centering
         \includegraphics[width=\textwidth]{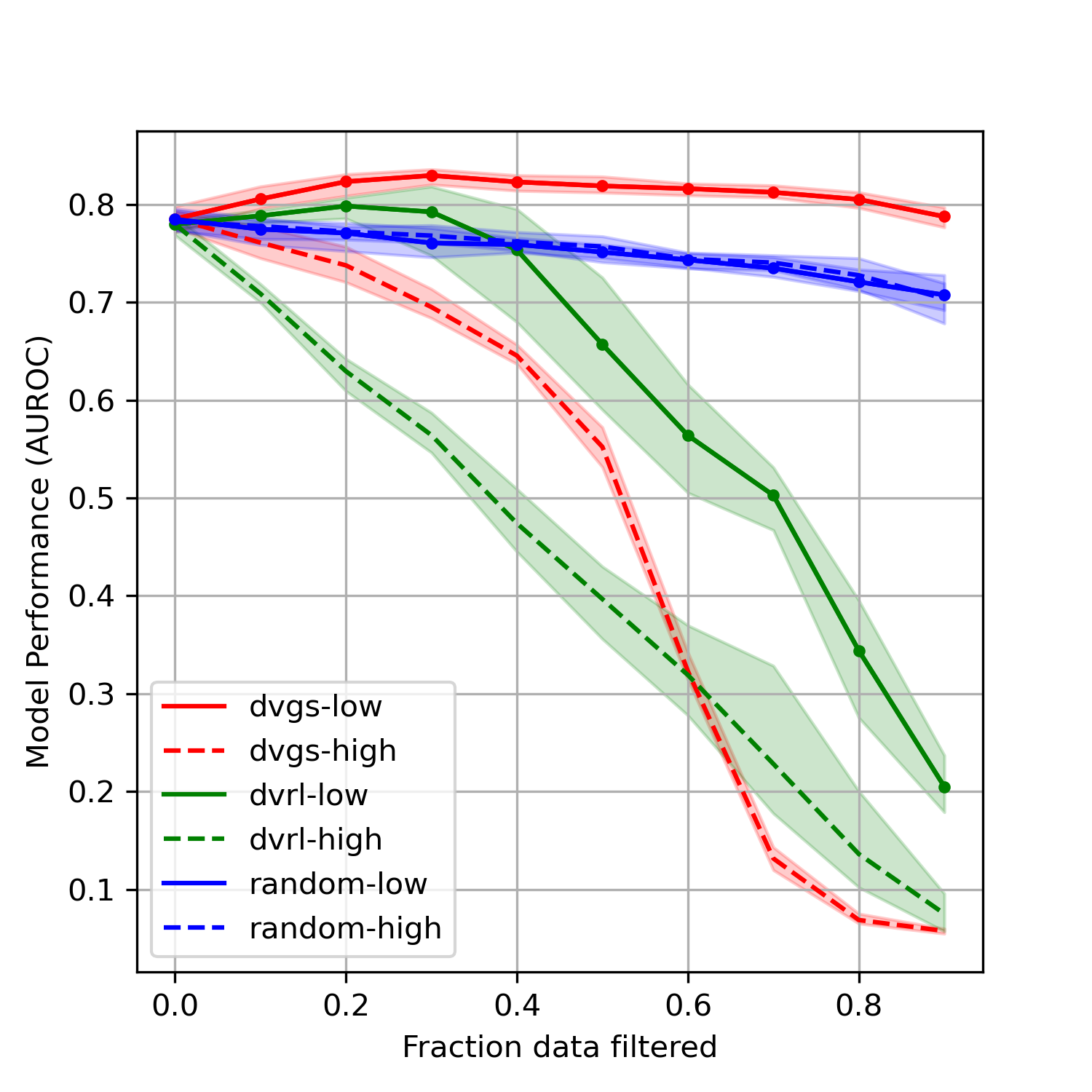}
         \caption{CIFAR10}
         \label{fig:exp3_perf}
     \end{subfigure}
\caption{The evaluation of respective data valuation methods ability to impact model performance when filtering either high value (dashed lines) or low values (solid lines).}
\label{fig:exp123_perf}
\end{figure*}

\begin{table*}[ht!]
\centering
\caption{The Area under the receiver operator curve (AUROC) scores if the data values are used to predict corrupted labels ($score = AUROC(noise\_labels, -data\_values)$); mean $\pm$ std.}
\label{table:exp123}
\begin{tabular}{|l|l|l|l|l|l|}
\hline
DATASET & DVGS              & DSHAP             & DVRL              & LOO               & RANDOM            \\ \hline
adult	& \textbf{0.896 $\pm$ 0.030} & 0.731 $\pm$ 0.049 & 0.887 $\pm$ 0.042 & 0.542 $\pm$ 0.056 & 0.503 $\pm$ 0.050 \\ \hline
blog	& \textbf{0.750 $\pm$ 0.028} & 0.671 $\pm$ 0.021 & 0.697 $\pm$ 0.033 & 0.558 $\pm$ 0.063 & 0.509 $\pm$ 0.028  \\ \hline
cifar10 & \textbf{0.954 $\pm$ 0.009} & NA                & 0.835 $\pm$ 0.110 & NA                & 0.499 $\pm$ 0.019 \\ \hline
\end{tabular}
\end{table*}

For all three datasets, we use a 2-layer neural network as the learning algorithm and the area under the receiver operator curve (AUROC) as the performance metric \cite{hanley1982meaning}. Each experiment is run at least five times with randomly sampled data subsets and unique weight initialization. Experiments are repeated to ensure stable results across diverse subsets of data and weight initialization.

Figure \ref{fig:exp123_labels} compares the ability of five data valuation methods to identify corrupt labels. Figure \ref{fig:exp123_perf} compares the effects of filtering based on data values on performance. In all three datasets, DVGS performs comparably or better than baseline data valuation methods. DVGS performs particularly well on the CIFAR10 dataset, which may be due to the informative features extracted from a pretrained InceptionNet model \cite{szegedy_inceptionNet}. 

The predictive quality of the data values for the identification of corrupt labels is shown in Table \ref{table:exp123}. DVGS data values are the most predictive of corrupted labels in all three datasets, as measured by the AUROC score. DVRL often performed comparably to DVGS, however, DVRL convergence was inconsistent and occasionally resulted in a suboptimal policy, as evidenced by the wide confidence intervals of DVRL in Figure \ref{fig:exp123_labels} and large standard deviations of CIFAR10 in Table \ref{table:exp123}. Additionally, we note that DVGS underperforms compared to Data Shapley when characterizing high data value, as seen in relative performance trends when filtering high-value data in Figure \ref{fig:exp123_perf}.

\FloatBarrier


\subsection{Characterization of Sample Noise}\label{sec:lincs_noise}

\begin{figure*}[ht!]
     \centering
     \begin{subfigure}{0.3\textwidth}
         \centering
         \includegraphics[width=\textwidth]{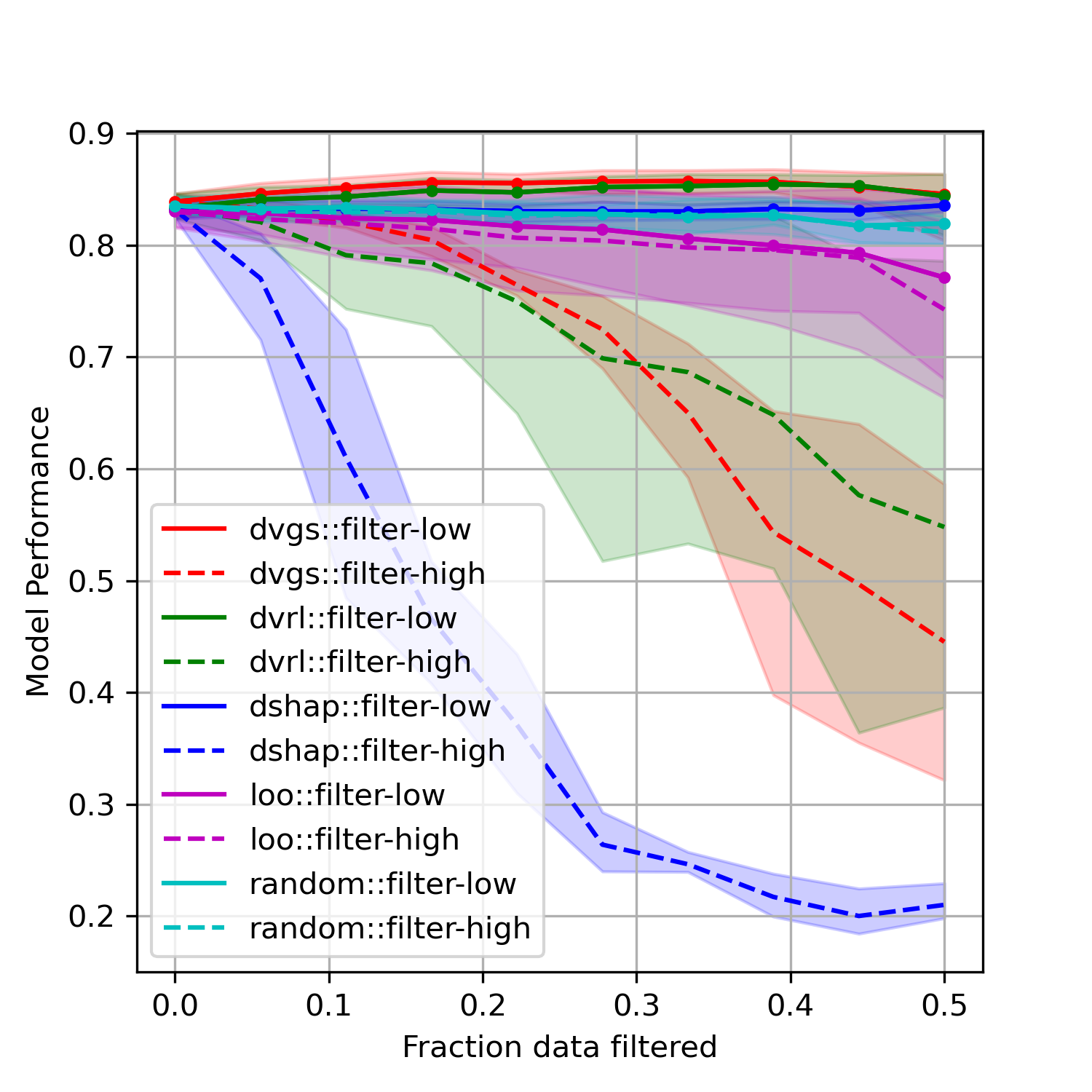}
         \caption{ADULT Dataset}
         \label{fig:exp4_perf}
     \end{subfigure}
     \hfill
     \begin{subfigure}{0.3\textwidth}
         \centering
         \includegraphics[width=\textwidth]{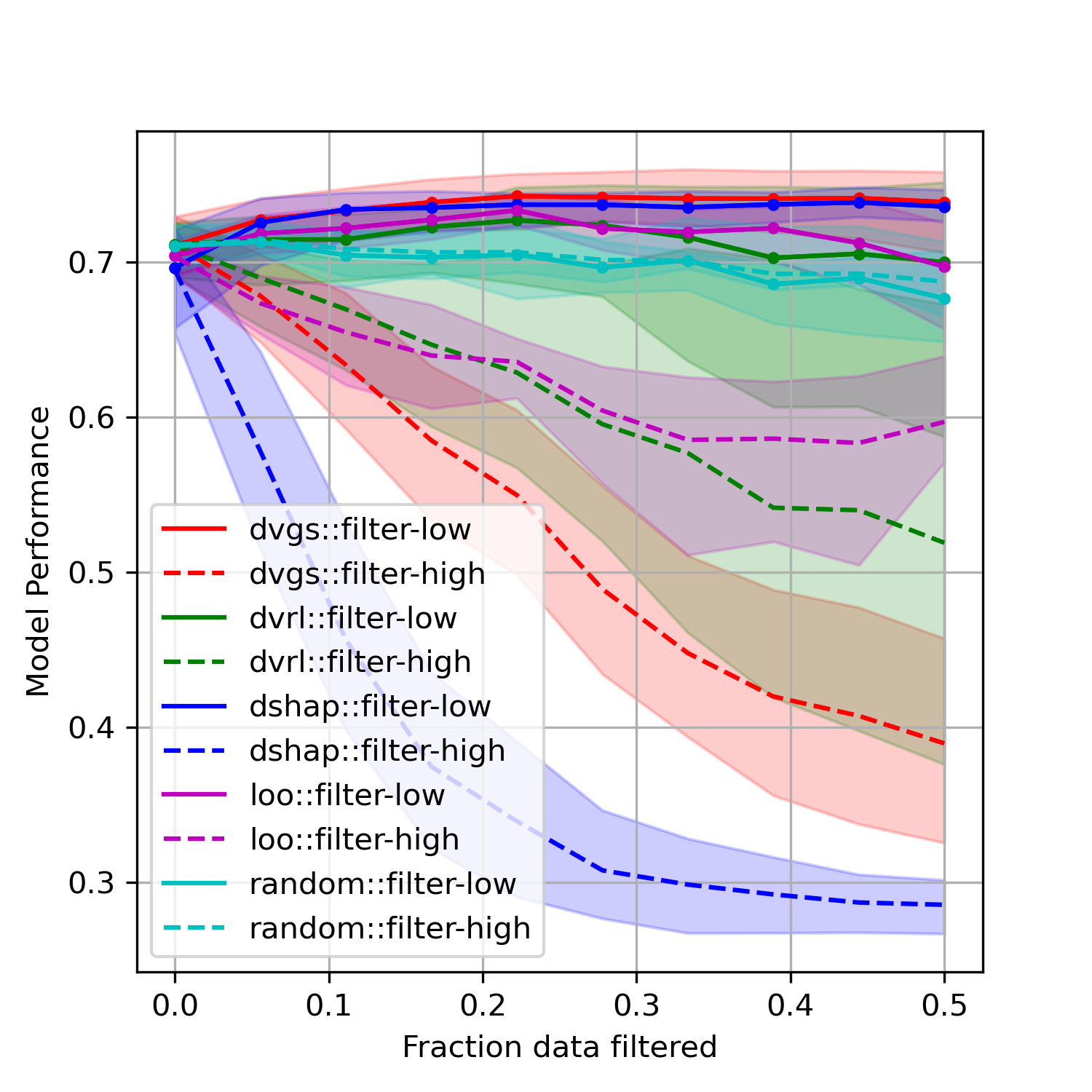}
         \caption{BLOG Dataset}
         \label{fig:exp5_perf}
     \end{subfigure}
     \hfill
     \begin{subfigure}{0.3\textwidth}
         \centering
         \includegraphics[width=\textwidth]{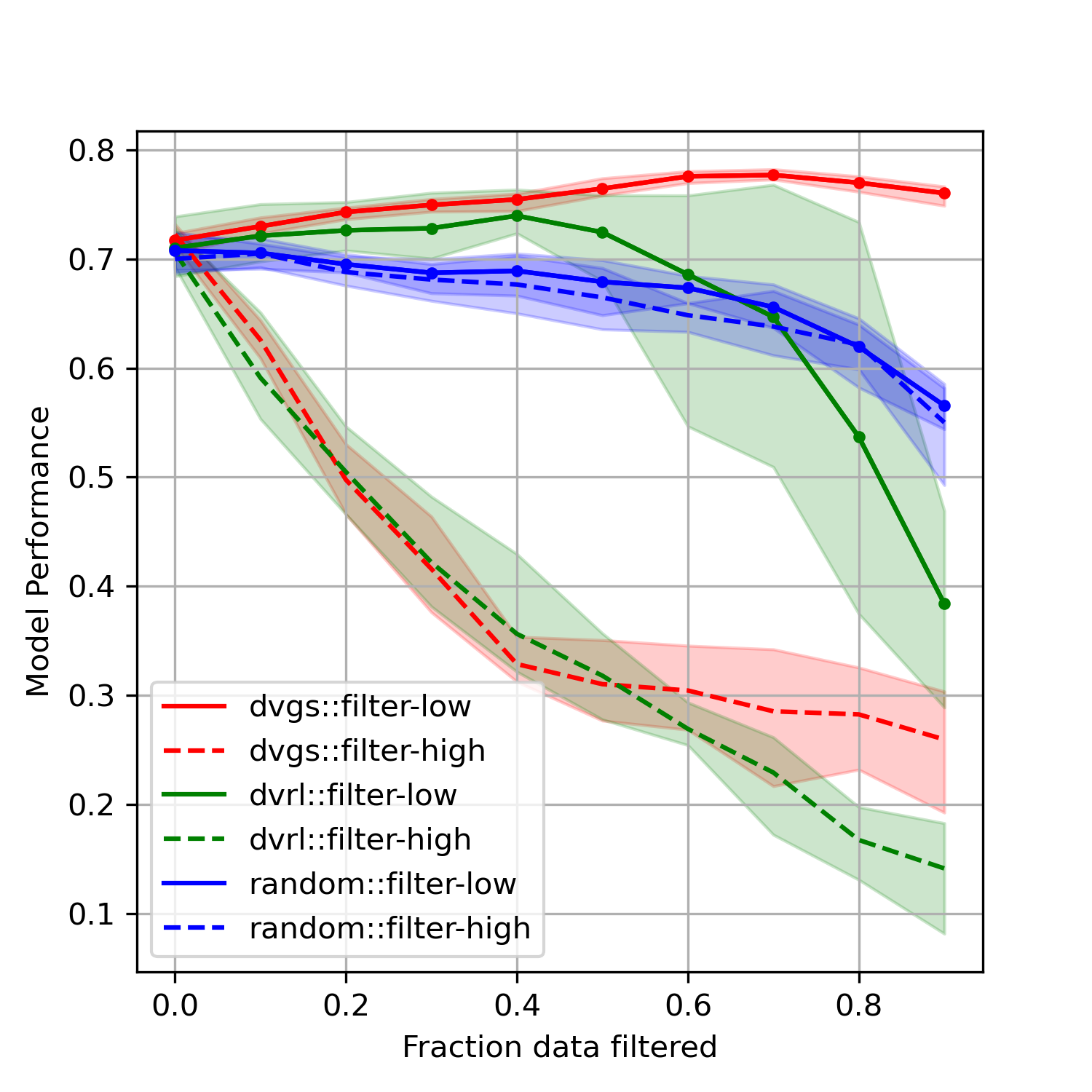}
         \caption{CIFAR10}
         \label{fig:exp6_perf}
     \end{subfigure}
        \caption{The evaluation of respective data valuation methods ability to impact model performance when filtering either high value (dashed lines) or low values (solid lines). The y-axis measures the model performance using the AUROC metric.}
\label{fig:exp456_perf}
\end{figure*}

In many domains, input features may be noisy due to measurement error, natural stochasticity, or batch effects, leading to inaccurate sample informativeness. To explore the ability of data valuation to quantify input feature noise, we artificially corrupt exogenous features as described in Section \ref{sec:methods}. For this task, we evaluate data valuation in supervised (ADULT, BLOG and CIFAR10) and unsupervised learning (CIFAR10 and LINCS) settings. In the supervised setting, we use architectures and hyper-parameters identical to those described in Section \ref{label corruption}. In unsupervised settings, we use an autoencoder architecture \cite{ae1, ae2} to create a low-dimensional representation and optimize using reconstruction mean square error (MSE). We justify that noisy samples will be more difficult to reconstruct and are likely to be detrimental to the performance. For the unsupervised setting, we apply our methods to two datasets: the CIFAR10 dataset and a high-quality subset of the LINCS L1000\footnote{Observations with an average Pearson correlation between replicates greater than 0.5}. The ability of the data values to characterize the exogenous feature noise rates is reported in Table \ref{table:exp456}. Compared to baseline methods, DVGS produces data values that most strongly correlate\footnote{More positive correlation is better performance; As described in Section \ref{sec:methods}, evaluation is performed by $Spearman(\phi_i, -\nu_i)$, since low data values are expected to correlate with high noise rates.} with ground-truth noise rates. As in Section \ref{label corruption}, we also evaluate the performance impact of filtering data based on data values, and these results are shown in Figure \ref{fig:exp456_perf}. We find that DVGS can most effectively characterize noise rates across all datasets. Additionally, when we compare model performance improvements when low value data are removed, as shown by the solid lines in Figure \ref{fig:exp456_perf}, we find that the performance of the DVGS method is comparable to or better than the baseline methods. As observed in the results of the supervised setting, we find that Data Shapley outperforms DVGS in quantifying high-quality data, measured by model performance decrease when filtering high-value data in both the ADULT and BLOG datasets, shown in Figure \ref{fig:exp456_perf} (a,b). In some of the learning tasks listed in Table \ref{table:exp456} only one or none of the baseline methods are calculated due to compute limitations. 

\begin{table*}[h!]
\centering
\caption{The Spearman correlation of predicted data values and artificial sample noise rates. The top performing method for each row is bolded; mean $\pm$ std.}
\label{table:exp456}
\resizebox{\textwidth}{!}{
\begin{tabular}{|r|r|r|r|r|r|r|}
\hline
\textbf{Dataset}             & \textbf{Learning}                 & \textbf{DVGS}      & \textbf{DSHAP}     & \textbf{DVRL}      & \textbf{LOO}       & \textbf{RANDOM}    \\ \hline
adult                        & supervised                        & \textbf{0.225 $\pm$ 0.061}& 0.130 $\pm$ 0.091& 0.159 $\pm$ 0.074& 0.022 $\pm$ 0.076& -0.007 $\pm$ 0.026\\ \hline
blog                         & supervised                        & \textbf{0.106 $\pm$ 0.077}& 0.086 $\pm$ 0.074& 0.100 $\pm$ 0.344& 0.045 $\pm$ 0.078& 0.011 $\pm$ 0.054\\ \hline
cifar10                      & supervised                        & \textbf{0.402 $\pm$ 0.081}& NA                 & 0.358 $\pm$ 0.103& NA                 & 0.000 $\pm$ 0.018\\ \hline
cifar10                      & unsupervised                      & \textbf{0.757 $\pm$ 0.131}& NA                 & NA                 & NA                 & 0.003 $\pm$ 0.014\\ \hline
lincs (APC\textgreater{}0.5) & \multicolumn{1}{r|}{unsupervised} & \textbf{0.505 $\pm$ 0.018}& NA                 & NA                 & NA                 & NA                 \\ \hline
\end{tabular}
}
\end{table*}


\subsection{Computational Complexity}

DVGS can be applied to large datasets and complex tasks with markedly lower computational costs than previous data valuation methods and enables application to new domains and data types. In Table \ref{table:runtime}, we show the runtime of four data valuation algorithms. On average, DVGS is roughly five times faster than DVRL and more than 100 times faster than truncated Monte-Carlo (TMC) Data Shapley. Compared to DVRL and Data Shapley, which require sequential training of models on different subsets of data, the DVGS method requires training only one model. Furthermore, by computing the gradient similarities every $T$ batches, the DVGS runtime can be reduced by a factor of $T$. In practice, we find that using values of $T$ between 2 and 5 has a marginal impact on the performance of the data values used for corrupted label discovery. These experiments are described in more detail in Supplementary Section \ref{supp:time}.

\begin{table*}[ht!]
\centering
\caption{Average runtime (in minutes) of 8 experiments. Experiments 1-3 were for label corruption; Experiments 4-6 were for noise characterization; Experiments 7 and 8 were unsupervised characterization of noise.}
\label{table:runtime}
\resizebox{0.6\textwidth}{!}{%
\begin{tabular}{|r|r|r|r|r|r|r|r|r|}
\hline
\textbf{method} & \textbf{exp1}     & \textbf{exp2}     & \textbf{exp3}     & \textbf{exp4}     & \textbf{exp5}     & \textbf{exp6}     & \textbf{exp7} & \textbf{exp8} \\ \hline
\textbf{dshap}  & 515.2        & 774.9        & NaN               & 404.5        & 631.0        & NaN               & NaN           & NaN           \\ \hline
\textbf{dvgs}   & \textbf{1.3} & \textbf{1.2} & \textbf{5.3} & \textbf{1.4} & \textbf{1.3} & \textbf{5.1} & 154.0    & 41.7      \\ \hline
\textbf{dvrl}   & 9.9          & 9.5          & 13.2         & 9.8          & 9.8          & 11.7         & NaN           & NaN           \\ \hline
\textbf{loo}    & 33.0         & 34.0         & NaN               & 35.1         & 34.7         & NaN               & NaN           & NaN           \\ \hline
\end{tabular}%
}

\end{table*}


\subsection{Data Valuation of the LINCS dataset}

In this section, we apply our DVGS method to quantify LINCS L1000 sample quality across all chemical perturbations. In each experiment, we randomly sampled a target and a test set (5000 observations each) in two conditions: 

\begin{itemize} 
    \item{\textbf{Noisy Target set} (high-APC). Target dataset sampled from all available observations.}
    \item{\textbf{Clean Target set} (all-APC). Target dataset sampled from high-APC observations (APC > 0.5).}
\end{itemize} 

In both configurations, we adjust the target set sampling probabilities so that the target set is balanced by perturbation type. The source set consists of all samples that are not in the target or test sets. See Supplementary \ref{supp:apc} for more information on APC calculation.   

Data valuation of LINCS could be done in a supervised or unsupervised setting, however, we chose to use an unsupervised prediction task for the following reasons: 

\begin{itemize} 

    \item{\textbf{Simplicity}: Encoding drug, cell line, concentration and measurement time requires additional overhead and may bias the results toward the encoding method chosen; e.g., encoded by drug targets, cell line expression, etc.}
    \item{\textbf{Imbalanced Dataset}: drug perturbations and cell lines are not equally represented in the LINCS dataset, and this may cause bias toward the over represented drugs or cell lines. While this is a concern in an unsupervised setting, we rationalize that removing exogenous variables may help mitigate the issue. Additionally, to further mitigate this concern we select a target set with more balanced proportions of drug perturbations.}
    \item{\textbf{Noise Quantification}: We consider measurement noise to be the primary data quality issue in the LINCS L1000 dataset and would like our data values to characterize sample noise rates. The results from Section \ref{sec:results} indicates that DVGS can effectively quantify sample noise using an unsupervised learning task.}

\end{itemize}

\begin{figure*}[ht!]
     \centering
     \begin{subfigure}{0.4\textwidth}
         \centering
         \includegraphics[width=\textwidth]{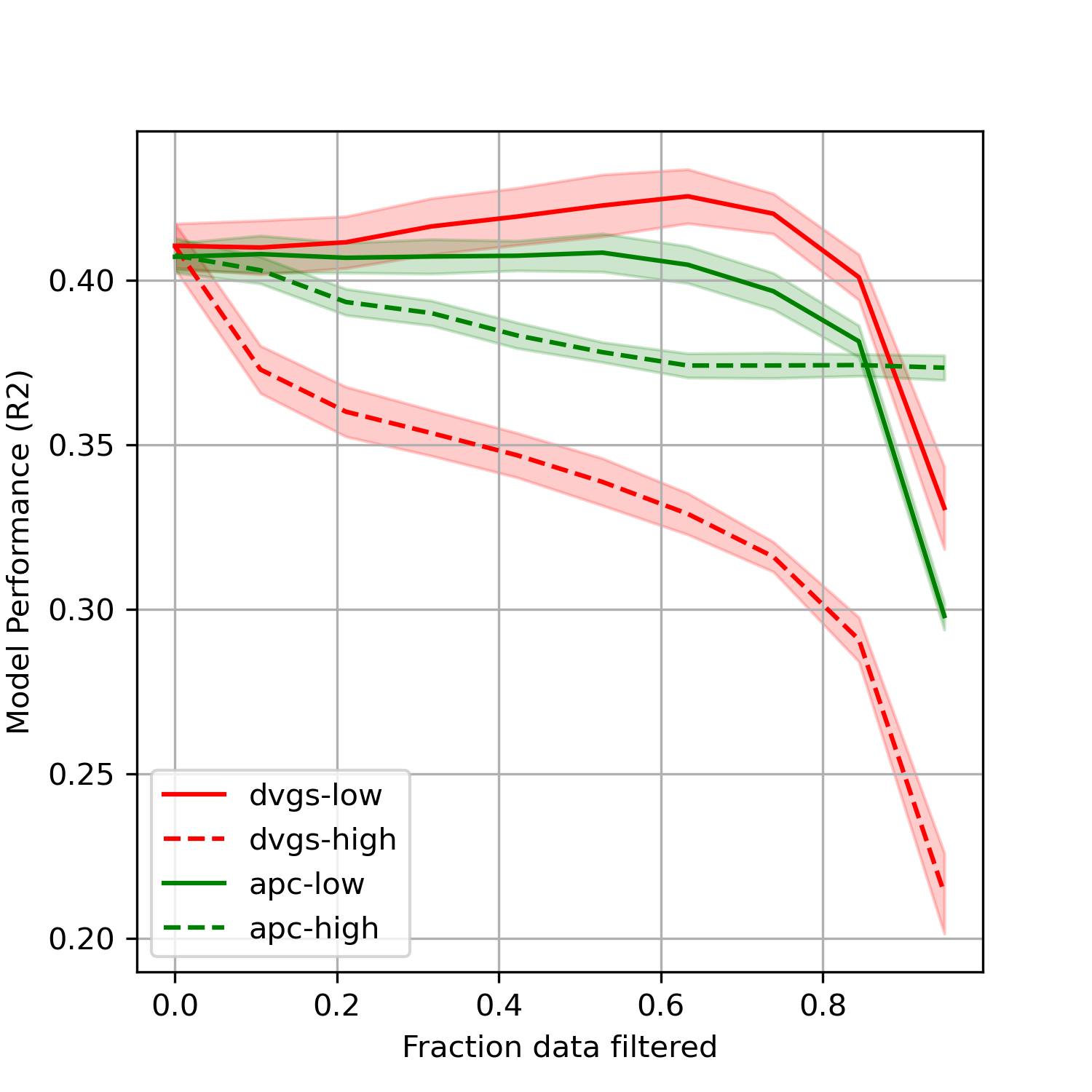}
         \caption{All-APC target set.}
         \label{fig:exp9_perf}
     \end{subfigure}
     \begin{subfigure}{0.4\textwidth}
         \centering
         \includegraphics[width=\textwidth]{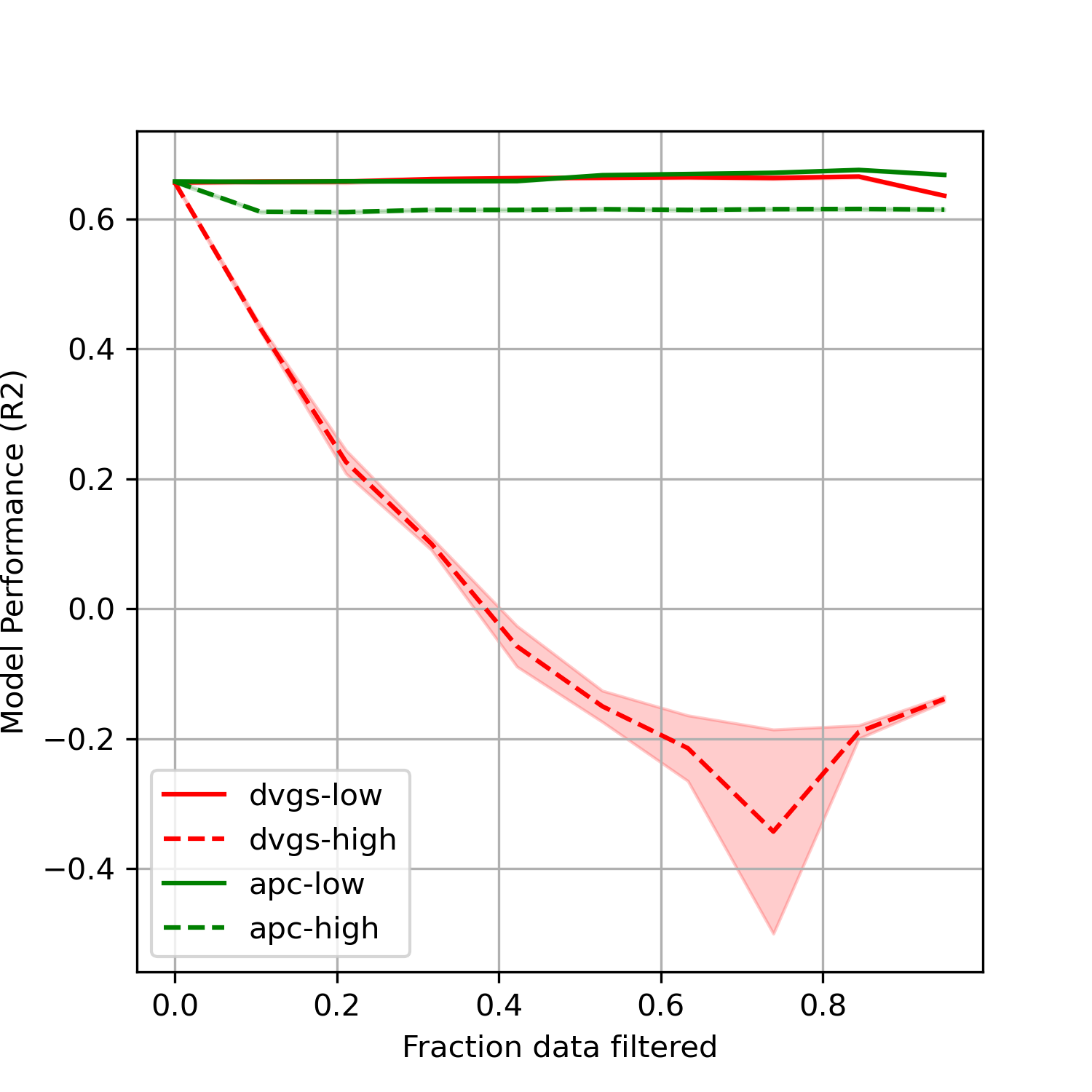}
         \caption{High-APC target set.}
         \label{fig:exp10_perf}
     \end{subfigure}
     \begin{subfigure}{0.4\textwidth}
         \centering
         \includegraphics[width=\textwidth]{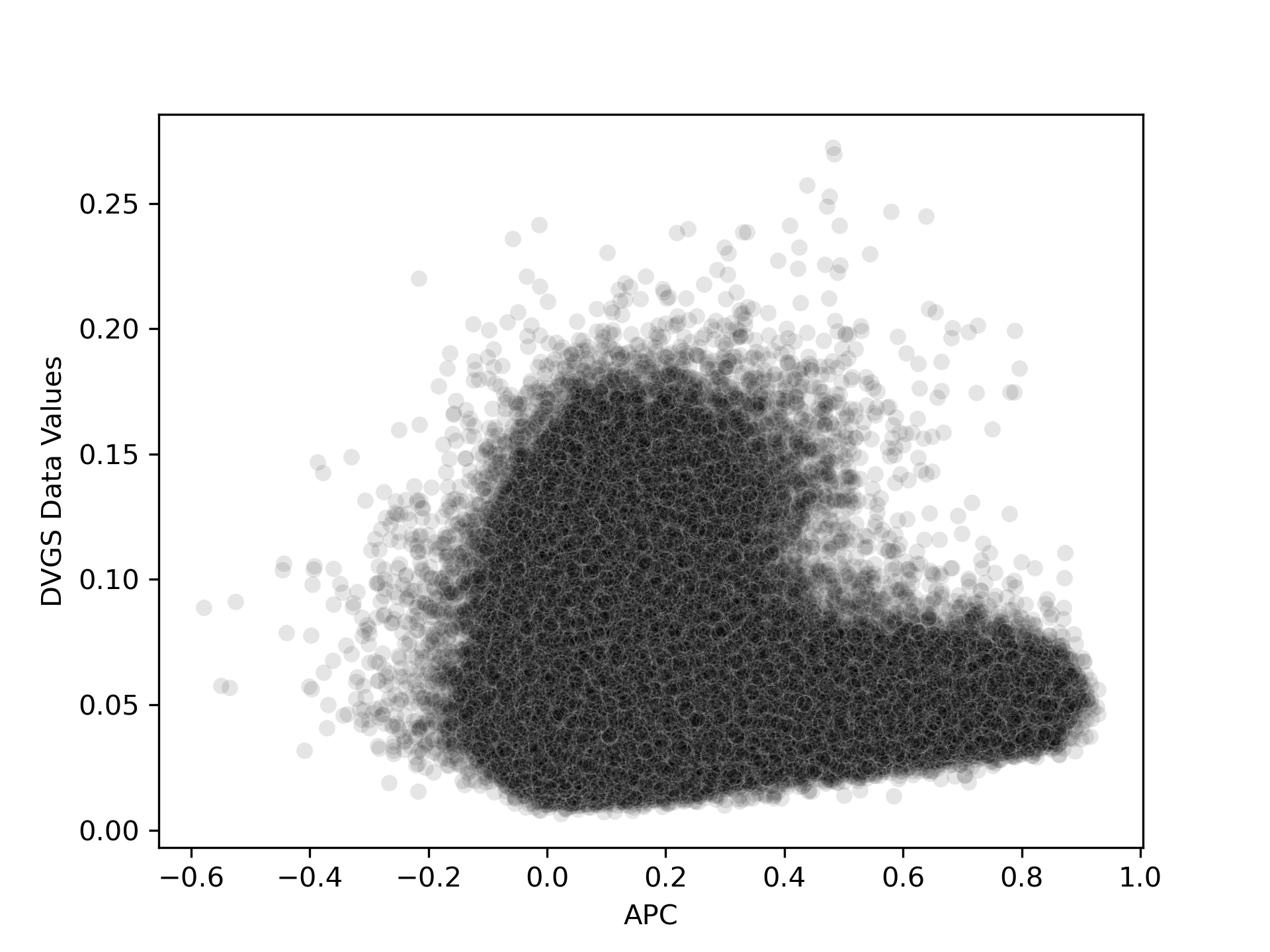}
         \caption{All-APC target set.}
         \label{fig:exp9_corr}
     \end{subfigure}
     \begin{subfigure}{0.4\textwidth}
         \centering
         \includegraphics[width=\textwidth]{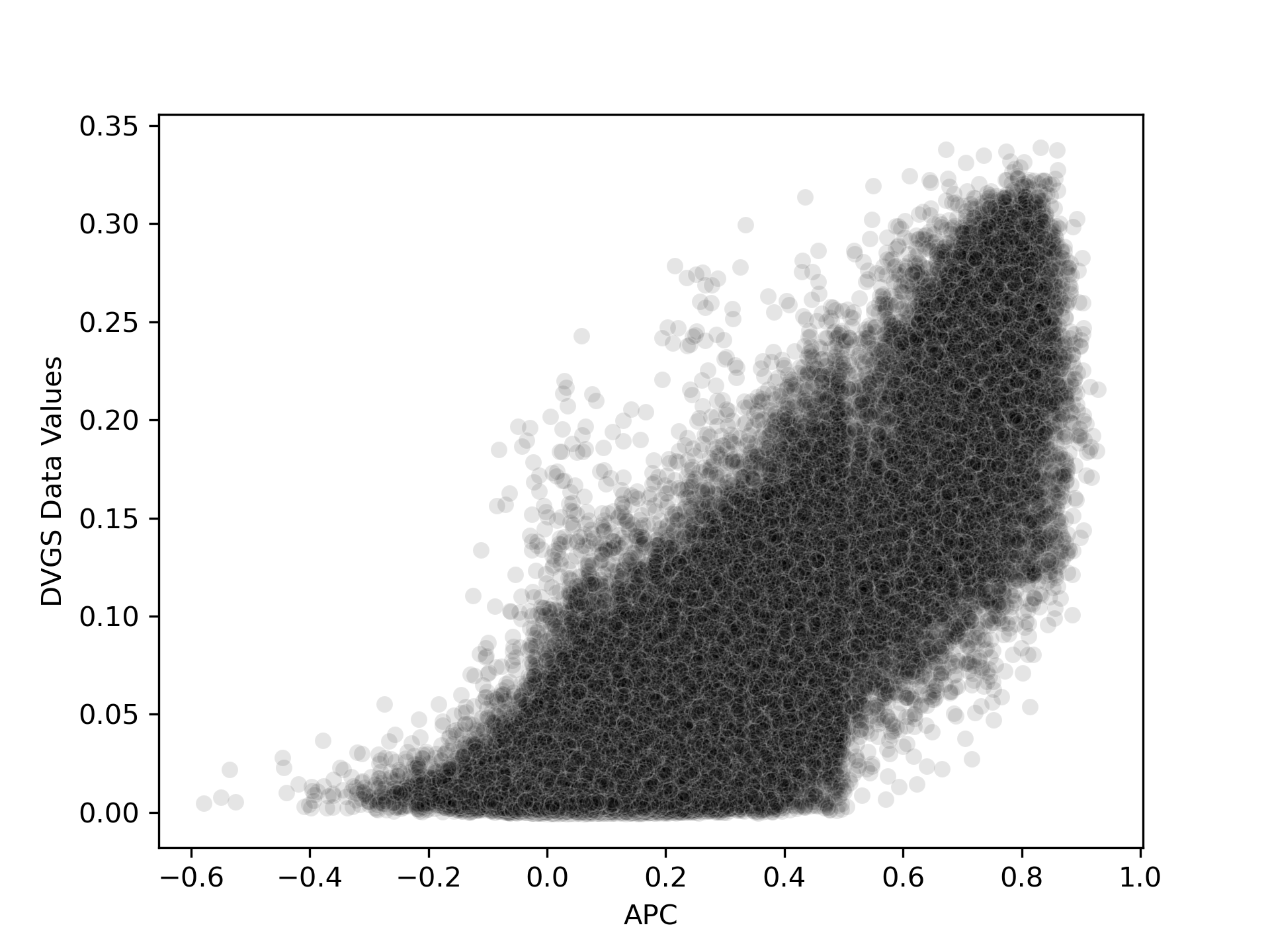}
         \caption{High-APC target set.}
         \label{fig:exp10_corr}
     \end{subfigure}
\caption{(a-b) The reconstruction performance ($R^2$) of autoencoders applied to the LINCS L1000 data when filtering low- and high- value data. (c-d) DVGS data values compared to APC values.}
\label{fig:exp9_10_perf}
\end{figure*}

For this task, we use an autoencoder with 2-layers in the encoder and decoder networks and 32 latent channels (embedding dimension). To avoid dependence on a specific target set, we ran the experiment several times ($n\geq3$) using different source, target, and test sets, as well as unique weight initializations. We compare the DVGS data values with the APC metric, proposed by Pham et al., to compare the generated data values to previous LINCS L1000 sample quality metrics. We evaluate the performance of LINCS data values by their ability to modify model performance when filtering high- and low-value data. Figure \ref{fig:exp9_10_perf} shows the performance comparison between the APC and DVGS data values. In the high-APC and all-APC conditions, we see that DVGS captures low data quality much better than the APC metric. In the all-APC condition, DVGS outperforms APC in capturing high-quality data, however, the DVGS data values and APC perform comparably in the high-APC condition. Additionally, we find that DVGS values and APC values correlate in the high-APC condition (Pearson Correlation $\sim$ 0.84) but not in the all-APC condition (Pearson Correlation $\sim$ -0.05). More specifically, in Figure \ref{fig:exp9_corr} we see that high APC values are depleted for high data values, suggesting that DVGS data values in the all-APC condition may characterize a different aspect of data quality or usefulness than APC. 

%% file: 05_discussion.tex
\section{Discussion}\label{sec:discussion}

In this work, we address scalability limitations of current data valuation methods by proposing a fast and robust method to estimate data values. We show that this method performs comparably or better than baseline methods in several tasks, including 1) identifying corrupted labels and 2) characterizing exogenous feature noise. Additionally, we have shown that our method works well to modify model performance when filtering data based on data values, and performs comparably or better than baselines when filtering low-value data. While Data Shapley and DVRL tend to lead to larger decreases in model performances when filtering high-value data, DVGS performs exceptionally well at identifying corrupted labels and noisy samples, especially in vision tasks using pretrained models. DVGS is also, on average, 100 times faster than Data Shapley (TMC) and 5 times faster than DVRL. This improvement in time complexity makes DVGS applicable to a wide range of datasets and domains. Additionally, in the reported experiments, DVGS was stable across hyperparameters (see Supplementary note \ref{supp:params}), data partition, and weight initialization. These characteristics make DVGS convenient and robust for many applications in data cleaning and machine learning. 

To show the value of our DVGS method in a real world scenario and to address data quality issues in a foundational dataset, we apply DVGS to the LINCS L1000 level 5 dataset that has more than 700k high-dimensional samples. We compare our method with a previous LINCS quality metric, the Average Pearson Correlation (APC), and show that our DVGS-produced data values are better able to modify model performance when filtering based on value. Interestingly, using a target dataset drawn randomly from the dataset (not necessarily high-quality) leads to data values that 1) do not correlate well with APC, and 2) significantly outperform APC as measured on a hold-out test set drawn from the full dataset. 

\subsection{Limitations and Future Directions}\label{sec:limitations}

Similarly to DVRL, our DVGS method lack the equitable data value properties proposed by Ghorbani et al., and therefore should not be interpreted in the same way; DVGS data values do not have a convenient interpretation like Data Shapley values. Rather, DVGS data values should be considered latent variables characterizing data usefulness, and we make no assumption about the linearity or magnitude of DVGS data values. These traits suggest that DVGS data values should be treated contextually as an ordered list of valuable samples. Pragmatically, ranked sample values meet the requirements of many of the evaluation techniques used by previous data valuation methods \cite{ghorbani_data_2019, yoon_data_2019} including identifying corrupted labels and noise quantification. Future directions may consider learning a task-specific function to estimate Data Shapley values from DVGS data values, which would allow users to interpret the DVGS data values in a way comparable to Data Shapley. This could be done by performing DVGS data valuation and calculating a limited number of Data Shapley values, which could then be used as a training set to infer Data Shapley values from DVGS values. Such an approach may help merge the scalability advantages of DVGS with the interpretability of Data Shapley. 

Through the lens of anomaly detection, DVGS can be viewed as a meta-learning algorithm that quantifies the similarity of the source samples to the target dataset and could potentially be used for anomaly detection. Additionally, this perspective may help explain why the DVGS method underperforms compared to baselines in identifying high-value data. For instance, if DVGS data values are considered a metric of similarity to the target set, then it may be that the most "similar" samples are not necessarily the most useful, whereas the most "dissimilar" data are likely erroneous or detrimental. It is therefore important that large data values be treated with caution. Additionally, it raises the question: how does DVGS handle redundant (or highly-similar) data in either the target or source datasets? Future work should address these concerns and characterize how redundancy can skew or alter DVGS data values. 

While DVGS works remarkably well on the evaluations listed in this paper, we do recognize that it is rare for gradient-based learning algorithms to be trained on gradient from single samples (e.g., on-line learning) and that most optimization algorithms are trained using mini-batches, thus implying that any sample's value or usefulness toward a predictive task cannot be considered independent of the other samples. Future work may wish to address this by looking at gradient similarity within mini-batches, or by selecting samples that align mini-batch gradients to the target dataset. One can imagine bi- or multi-modal sample-gradients, all of which may align poorly to a target mini-batch gradient, but when source samples are averaged in a mini-batch may align far more closely.  

%% file: 06_supplementary.tex
\section{Supplementary}\label{sec:supplementary}

\subsection{DVGS Robustness to Hyperparameters} \label{supp:params} 

\begin{figure}[h]
\centering
\includegraphics[width=1\linewidth]{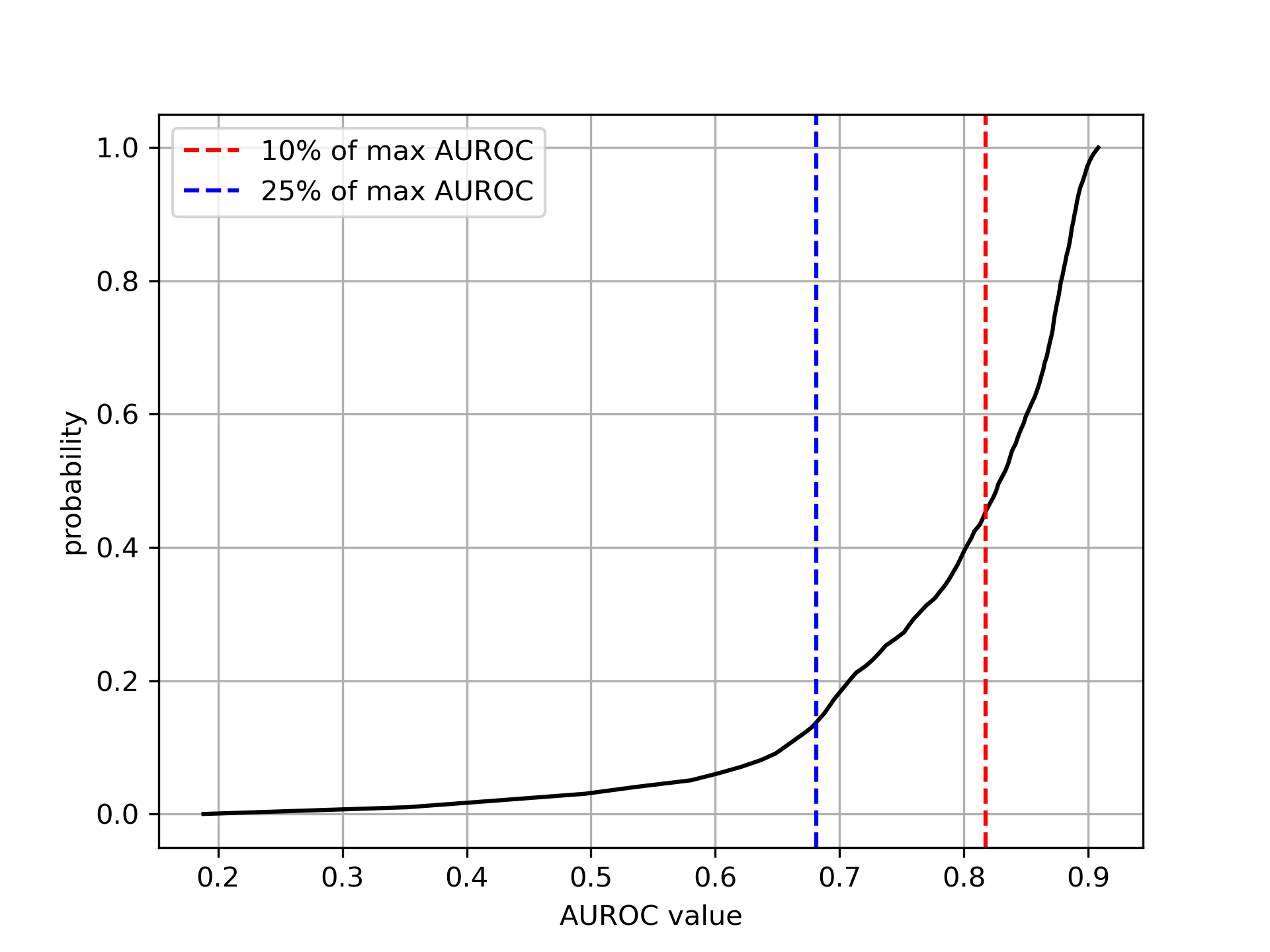}
\caption{The cumulative distribution function (CDF) of $AUROC(c_i, -\nu_i)$ across all tested hyperparameters, where $\nu_i$ are data values generated by DVGS and $c_i$ are the corrupted labels label. The red dashed line demarcates all AUROC values larger than this are within 10\% of the max AUROC value (e.g., roughly 55\% of all tested hyperparameters resulted in an AUROC value within 10\% of the max AUROC).}
\label{fig:hyperparam_cdf}
\end{figure}

To test the robustness of the DVGS method with respect to algorithm hyperparameters, we performed a grid search on the ADULT dataset with 20\% corrupted endogenous labels. We record the ability of DVGS to identify the corrupted labels across all tested hyperparameters. Figure \ref{fig:hyperparam_cdf} shows the cumulative distribution function (CDF) of the resulting AUROC values across all hyperparameters tested. Note that the AUROC metric characterizes the ability of data values to classify corrupted labels. We find that almost 85\% of the tested hyperparameter configurations resulted in performances within 25\% of the maximum performance, and more that 50\% of the tested hyperparameters resulted in performance within 10\% of the maximum performance, indicating that the DVGS method is robust to choice of hyperparameters. The hyperparameter grid search configurations are shown in Table \ref{table:hyperparam_options}. 

\begin{table*}[h!]
\begin{tabular}{|l|l|l|}
\hline
Hyperparameter         & Values                                                       & Optimal value\\ \hline
balanced class weights & True, False                                                  & False                                                                  \\ \hline
dropout                & 0, 0.25, 0.5, 0.75                                           & 0.25                                                                   \\ \hline
target batch size      & 100, 200, 400                                                & 200                                                                    \\ \hline
similarity             & Euclidean, Cosine Similarity, Dot Product, Scalar Projection & Euclidean                                                              \\ \hline
learning rate & 1e-2, 1e-3, 1e-4                                             & 1e-3                                                                   \\ \hline
Instance normalization& True, False                                                  & True                                                                   \\ \hline
number of layers& 1,2                                                          & 1                                                                      \\ \hline
activation function& Mish, ReLU                                                   & Mish                                                                   \\ \hline
\end{tabular}
\caption{The DVGS hyperparameter configurations tested in a grid search with 2 replicates per configuration.}
\label{table:hyperparam_options}
\end{table*}

\subsection{Average Pearson Correlation (APC) metric} \label{supp:apc} 

We compute the previously proposed Average Pearson Correlation (APC) \cite{pham_deep_2021} of LINCS level 4 replicates using the procedure: 

For a given level 5 LINCS sample: 

\begin{itemize} 

    \item{Identify the level 4 bio-replicate \textit{sample id}s that were used to generate the level 5 aggregate sample.} 
    \item{Load the level 4 sample ID expression profile into memory} 
    \item{Filter to select only landmark genes (978)} 
    \item{Compute the average pairwise Pearson correlation of level 4 bio-replicates}

\end{itemize} 

\begin{figure}[h]
\centering
\includegraphics[width=1\linewidth]{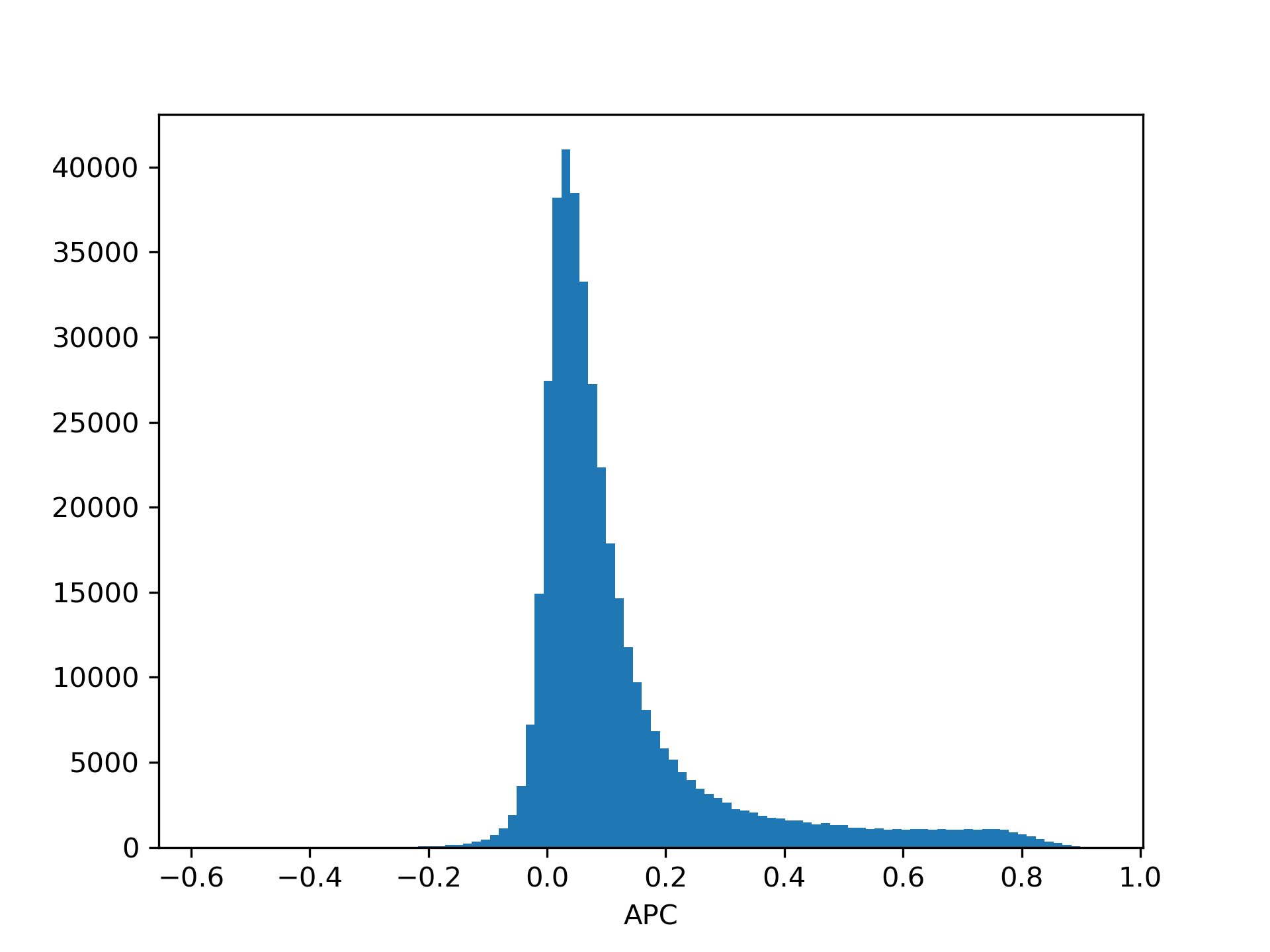}
\caption{The Average Pearson Correlation (APC) distribution of level 5 LINCS samples.}
\label{fig:apc_dist}
\end{figure}

As shown in Figure \ref{fig:apc_dist}, the resulting APC distribution is skewed right, with the majority of samples having an APC less than 0.5, suggesting that most of the replicates are highly discordant. Notably, future work may wish to perform data valuation directly on the level 4 samples, which may enable researchers to "rescue" high-quality replicates, even if the replicates are highly discordant. 

\FloatBarrier

\subsection{Additional Runtime Experiments} \label{supp:time} 

\begin{figure*}[h!]
     \centering
     \begin{subfigure}{0.45\textwidth}
         \includegraphics[width=\textwidth]{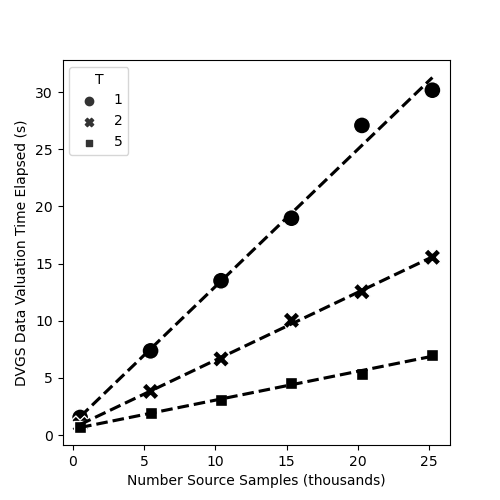}
         \caption{DVGS runtime on the ADULT dataset when computing gradient similarities every T steps.}
         \label{fig:scalability_time}
     \end{subfigure}
     \hfill
     \begin{subfigure}{0.45\textwidth}
         \centering
         \includegraphics[width=\textwidth]{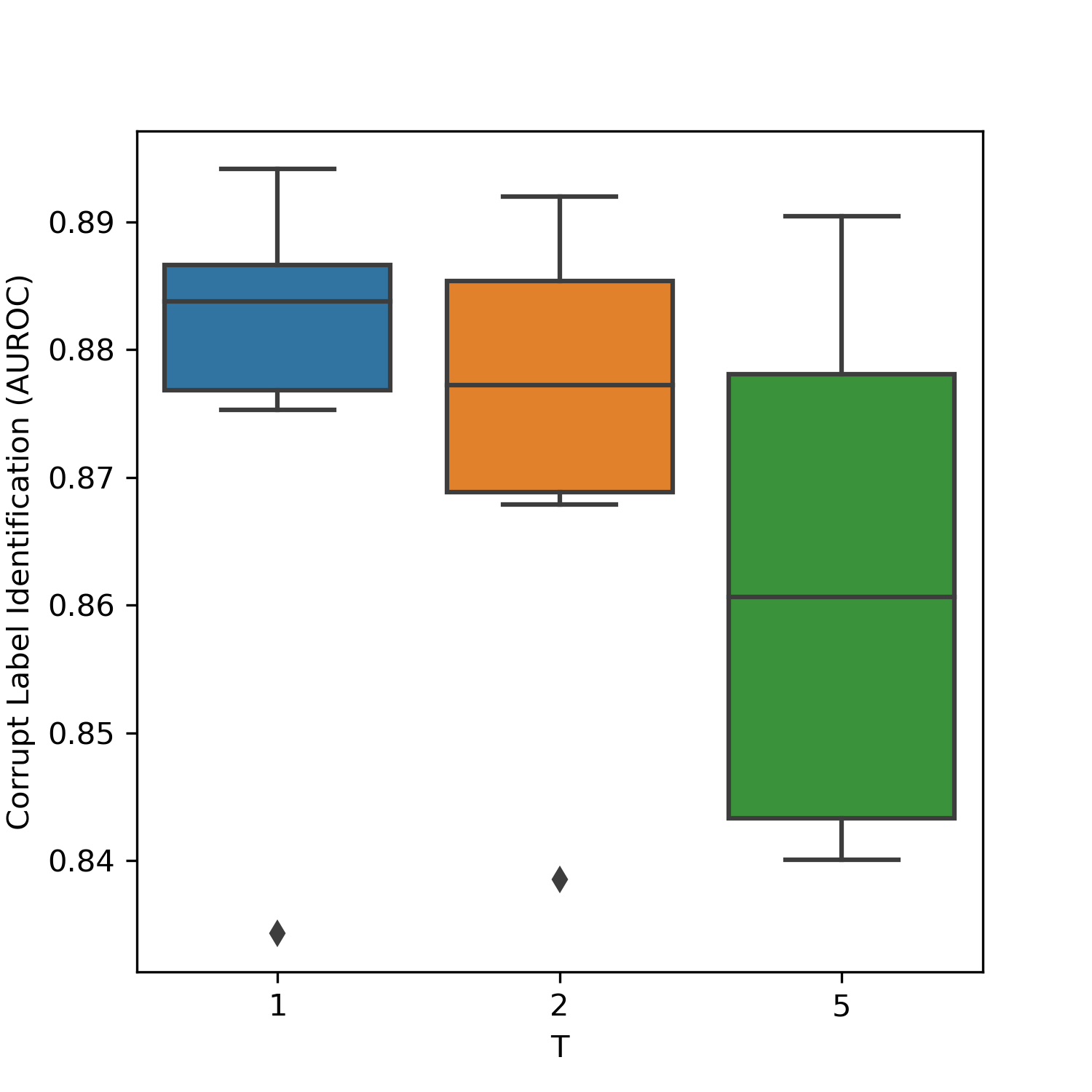}
         \caption{Ability of DVGS to identify corrupted labels, with different values of T (period of source gradient computations).}
         \label{fig:scalability_perf}
     \end{subfigure}
     \hfill
\caption{The scalability and performance of the DVGS method dependant on number of source samples and the period of source similarity computations (T).}
\label{fig:scalability}
\end{figure*}

In Figure \ref{fig:scalability} we show the experimental results of DVGS as the number of source samples increases. As expected, DVGS scales linearly with the number of source samples, divided by the period of gradient computations ($T$). In Figure \ref{fig:scalability_perf} we show the ability of DVGS to classify corrupted labels, when we increase the value of $T$, as one would expect, the AUROC value decreases with larger T, however, the marginal decrease in performance may be worthwhile for the improvements in runtime, especially on large datasets. When applying our method to the LINCS dataset, we were able to run 500 epochs of DVGS on 710,216 source samples using a multilayer autoencoder neural network (Number parameters > 650k) in roughly 8 hours on a Nvidia 3090 GPU. 

The memory requirement of the DVGS method is in many ways comparable to classical SGD optimization problems; however, the computation of high-dimensional sample-wise gradients can increase the memory requirements. Therefore, as the number of model parameters increases, the memory footprint of the sample gradients will also increase. To mitigate this issue, we chose to compute sample gradients in mini-batches, which can be manually specified to fit a given task. Reducing the source batch size will therefore reduce the memory footprint, but lead to a small increase in computation time. Additionally, the user can also choose to select a subset of all the model parameters to use for gradient computation, which will reduce memory overhead. 

